\documentclass{article}

\usepackage[main, final]{neurips_2026}

\usepackage[utf8]{inputenc} 
\usepackage[T1]{fontenc}    
\usepackage[
    colorlinks=true,
    linkcolor=blue!50!black,
    citecolor=blue!50!black,
    urlcolor=blue!50!black
]{hyperref}
\usepackage{url}            
\usepackage{booktabs}       
\usepackage{amsfonts}       
\usepackage{nicefrac}       
\usepackage{microtype}      
\usepackage{xcolor}         
\usepackage{subcaption}

\usepackage{multirow}
\usepackage{enumitem}
\usepackage{wrapfig}
\usepackage{bm}
\usepackage{algorithm}
\usepackage{algorithmic}

\usepackage{amsmath}
\usepackage{amssymb}
\usepackage{mathtools}
\usepackage{amsthm}

\usepackage[capitalize,noabbrev]{cleveref}

\usepackage{tcolorbox}
\usepackage{float} 

\tcbset{
  takeawaybox/.style={
    colback=gray!10!white,
    colframe=gray!60!black,
    fonttitle=\bfseries\small,
    title=Takeaway,
    boxrule=0.5pt,
    arc=3pt,
    left=4pt, right=4pt, top=2pt, bottom=2pt,
  }
}

\theoremstyle{plain}

\theoremstyle{definition}

\theoremstyle{remark}

\usepackage{placeins}

\definecolor{rliableolive}{HTML}{BBCC33}
\definecolor{rliableblue}{HTML}{77AADD}
\definecolor{rliablered}{HTML}{EE8866}

\tcbset{
  aibox/.style={
    width=\linewidth,
    top=8pt,
    bottom=4pt,
    colback=gray!7!white,
    colframe=black,
    colbacktitle=black,
    enhanced,
    center,
    attach boxed title to top left={yshift=-0.1in,xshift=0.15in},
    boxed title style={boxrule=0pt,colframe=white,},
  }
}
\newtcolorbox{AIbox}[2][]{aibox,title=#2,#1}

\title{Reward-Conditioned Reinforcement Learning}

\author{%
  Michal Nauman \\
  University of Warsaw\\
  \texttt{nauman.mic@gmail.com} \\
  \And
  Marek Cygan \\
  University of Warsaw, Nomagic \\
  \And
  Pieter Abbeel \\
  UC Berkeley, Amazon FAR\\
}

\begin{document}

\maketitle

\begin{abstract}

Sample-efficient reinforcement learning typically uses a fixed reward function both to guide exploration and to supervise learning. However, the same experience can provide supervision under multiple related reward specifications, even when the objective of interest remains fixed. We introduce \emph{Reward-Conditioned Reinforcement Learning} (RCRL), an off-policy approach that collects experience under a nominal objective while training a reward-conditioned agent using counterfactual rewards recomputed from replay. RCRL constructs alternative reward specifications by perturbing the nominal reward, providing diverse supervision without additional environment interaction. Across single-task, multi-task, and vision-based control benchmarks, RCRL improves nominal task performance and sample efficiency when integrated with strong RL backbones. As a secondary capability, the learned policies support zero-shot behavioral adjustment toward related objectives without further training. These results demonstrate that diverse reward supervision can improve learning of a predefined objective while also enabling policy steerability.

\end{abstract}

\section{Introduction}

Training on diverse tasks and objectives is a common strategy for improving representations and performance in machine learning~\citep{achiam2023gpt, guo2025deepseek}. In reinforcement learning (RL), however, agents trained for a known task typically use a single reward to both guide exploration and supervise learning~\citep{fujimoto2018addressing,haarnoja2018soft}. Yet choosing one objective for interaction does not require learning from only one reward. To this end, we ask whether diverse reward supervision can improve learning on a predefined task of interest.

Many control problems use rewards that combine task progress, tracking, and control costs~\citep{tassa2018deepmind,peng2018deepmimic} and thus provide an opportunity to broaden supervision without collecting additional experience. Standard RL combines these components into a scalar reward and uses only that reward to supervise learning~\citep{sutton2018reinforcement}. But these components can be recombined, and the same transition can be evaluated under many objectives. Such counterfactual rewards can provide alternative learning targets from the same experience. However, it is unclear if such diverse reward supervision improves the original task. 

We study this question through \textbf{Reward-Conditioned Reinforcement Learning (RCRL)}. RCRL relabels replayed transitions with alternative reward specifications and conditions the agent on the specifications used for each update. We construct these either by perturbing the nominal rewards or by reusing reward functions from the related tasks. Because rewards are recomputed from components stored in replay, RCRL adds diverse supervision without additional environment interactions, while 

\begin{figure}[H]
  \centering
  \vspace{-6pt}
  \begin{subfigure}{0.24\linewidth}
    \centering
    \includegraphics[width=0.95\linewidth]{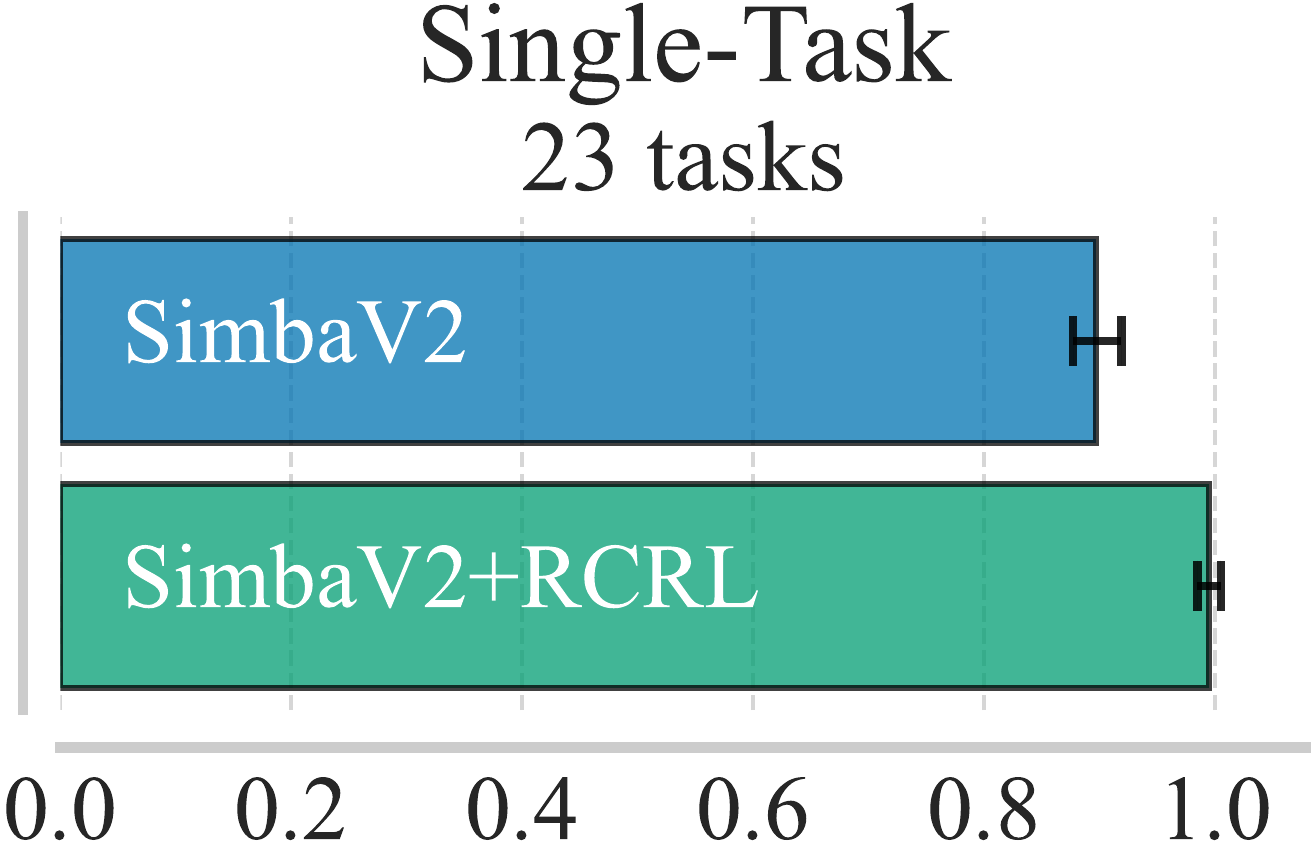}
  \end{subfigure}\hfill
  \begin{subfigure}{0.24\linewidth}
    \centering
    \includegraphics[width=0.95\linewidth]{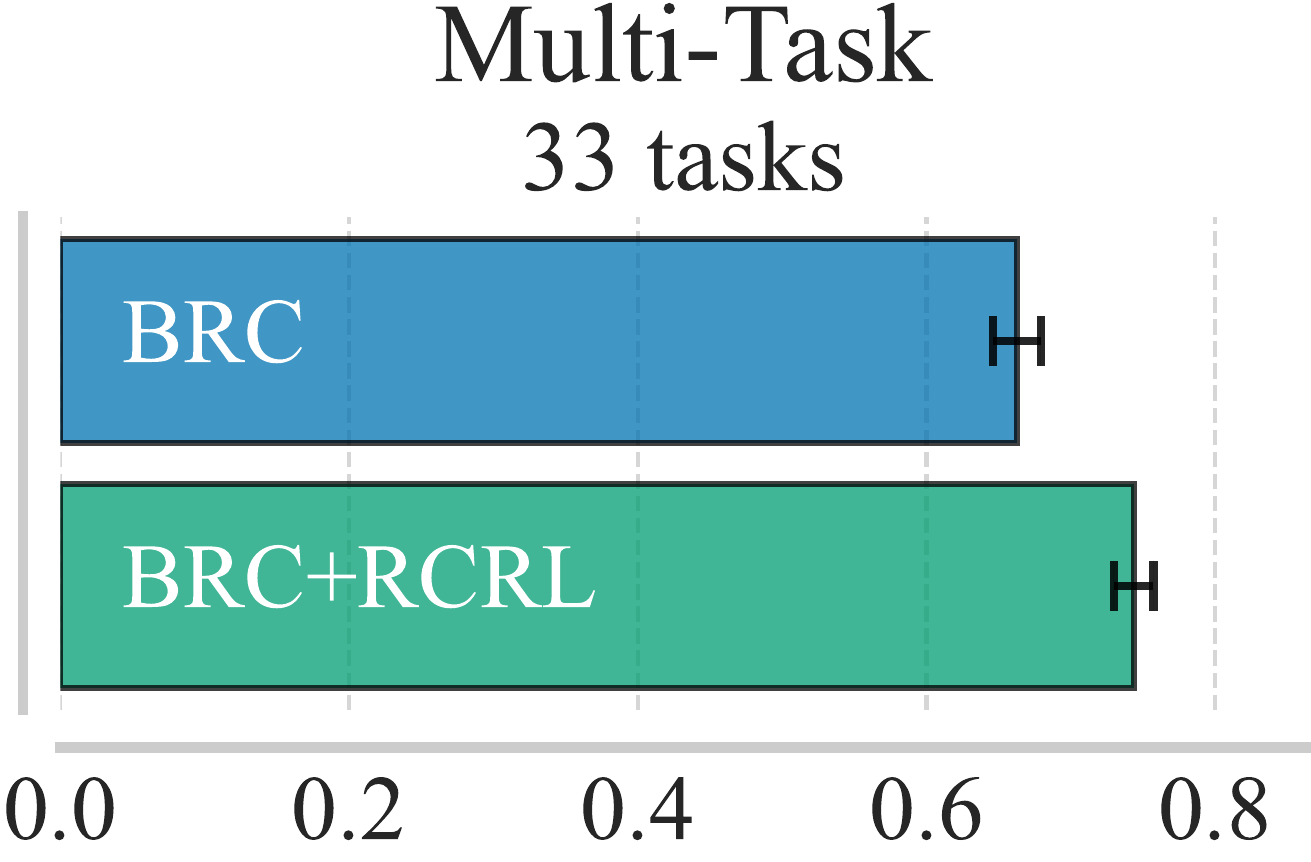}
  \end{subfigure}\hfill
  \begin{subfigure}{0.24\linewidth}
    \centering
    \includegraphics[width=0.95\linewidth]{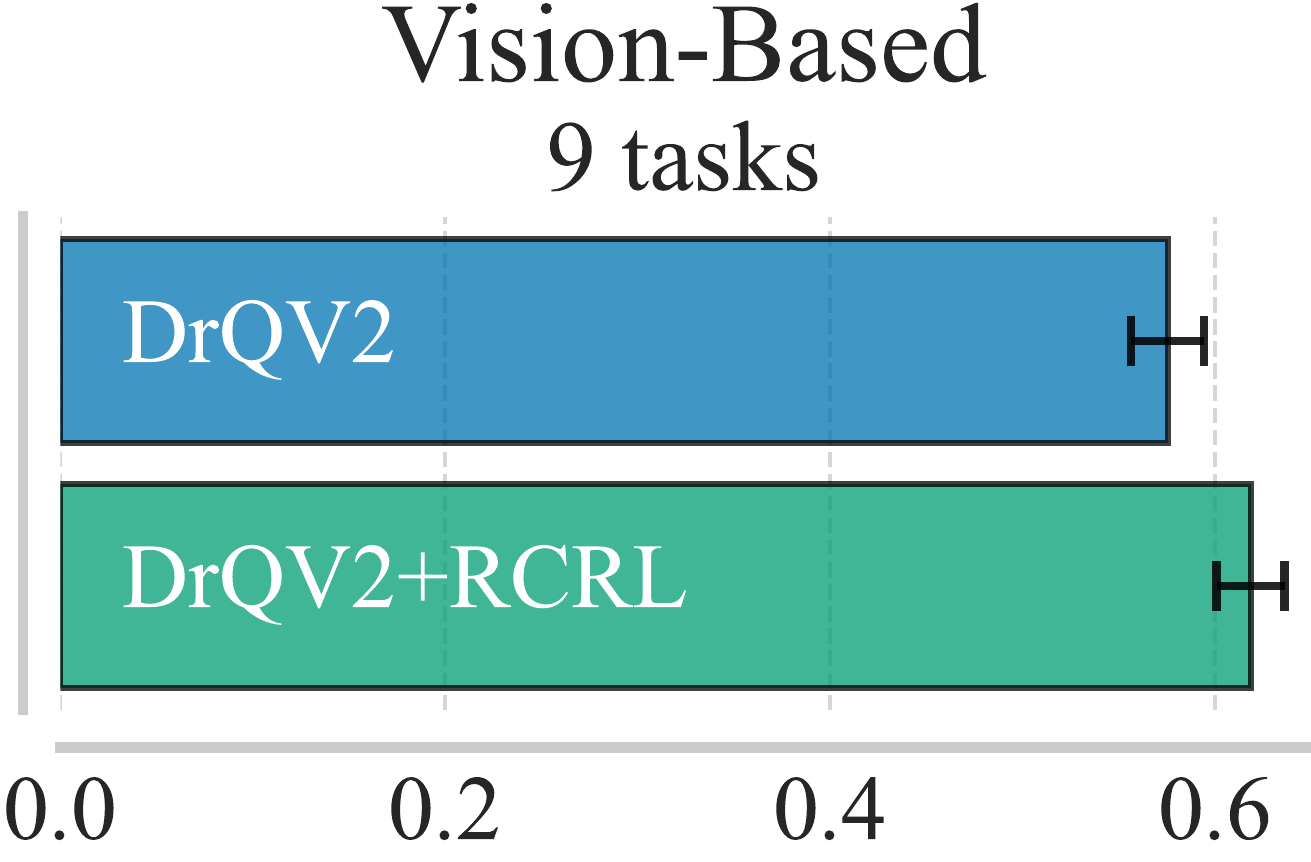}
  \end{subfigure}\hfill
  \begin{subfigure}{0.24\linewidth}
    \centering
    \includegraphics[width=0.95\linewidth]{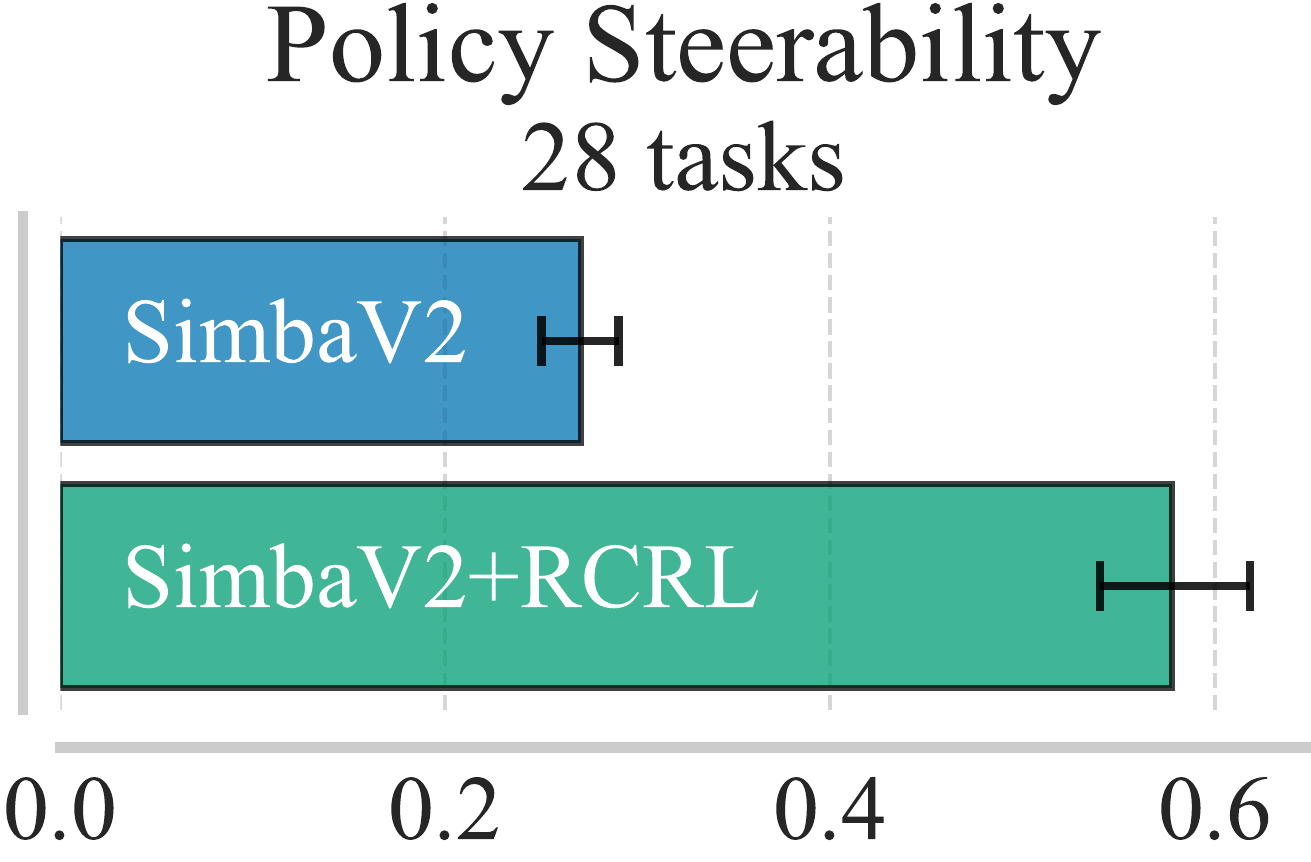}
  \end{subfigure}
  \vspace{-5pt}
  \caption{\footnotesize
    \emph{\textbf{Results summary.}}
    RCRL improves the performance of strong baselines
    in both single and multi-task settings
    (Figure~\ref{fig:iclr_timeseries}).
    As a secondary capability, RCRL increases
    the steerability of the trained policy without
    further training (Figure~\ref{fig:results_zeroshot}).
  }
  \label{fig:aggregate_results_small}
  \vspace{-6pt}
\end{figure}
\begin{wrapfigure}[14]{r}{0.48\textwidth}
  \centering
  \vspace{-10pt}
  \includegraphics[width=0.425\textwidth]{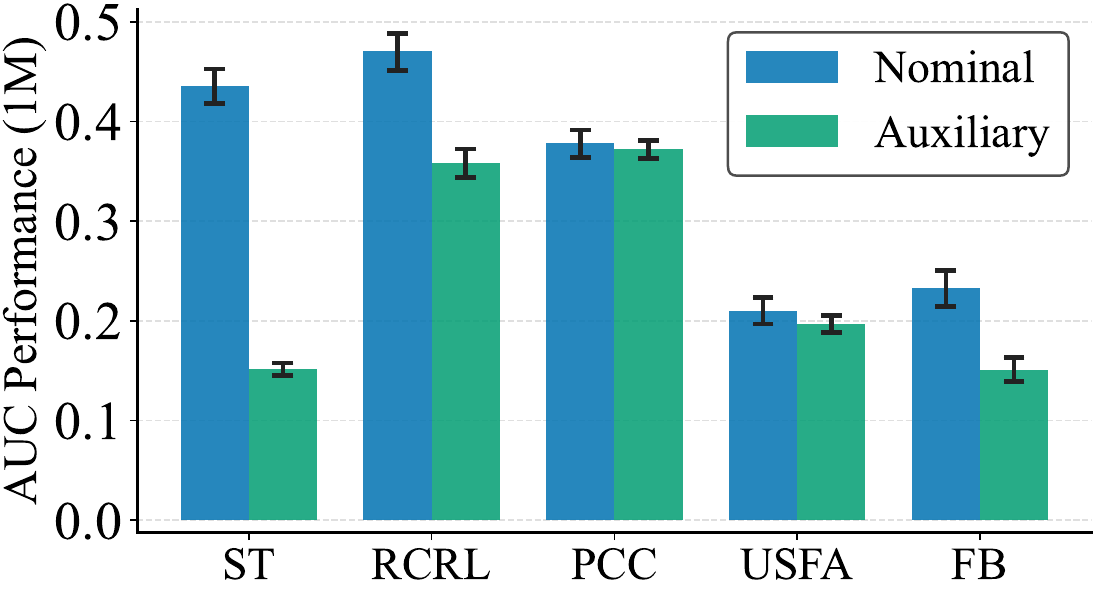}
  \vspace{-2pt}
  \caption{\footnotesize \emph{\textbf{Multi-objective and reward-free methods.}} Despite exploring only under the nominal task, RCRL achieves the strongest nominal performance while allowing for steerability which is comparable to multi-objective PCC.}
  \label{fig:auxiliary_ablation1}
\end{wrapfigure}


exploration focuses fixed \emph{nominal objective}. We evaluate RCRL with strong backbones across single-task, multi-task, and vision-based control, including \textsc{SimbaV2}~\citep{lee2025hyperspherical}, \textsc{BRC}~\citep{nauman2025bigger}, and \textsc{DrQV2}~\citep{yarats2021drqv2}. Across these, RCRL improves nominal task sample efficiency, while allowing for steering toward related objectives without further training. This shows that diverse reward supervision can improve learning of the target objective while making the policy more steerable. Our findings are as follows:

\begin{itemize}[leftmargin=10pt, itemsep=1pt, topsep=0pt]

\item \textbf{Improved Sample Efficiency} --
Training with diverse reward supervision via RCRL improves sample efficiency of strong RL backbones, despite evaluation being focused only on the nominal task. 
(Figures~\ref{fig:aggregate_results_small},~\ref{fig:iclr_timeseries} \& \ref{fig:dogs_results}).

\item \textbf{Improved Steerability} --
As a secondary capability, RCRL enables a trained policy to adjust its behavior without further training. Diverse exploration strengthens this steerability at cost to nominal performance (Figures~\ref{fig:results_zeroshot} \& \ref{fig:auxiliary_ablation1}).

\end{itemize}

\section{Related Work}
\label{sec:rw}

In RCRL, environment interactions follow a predefined objective, while diverse rewards are used for supervision. This setup is motivated by improving learning of the predefined task, while also allowing for steerability within the reward family. Table below summarizes the main differences between the proposed and related approaches. We expand the related work discussion in Appendix~\ref{app:related_works}.

\begin{table*}[ht]
\centering
\small
\setlength{\tabcolsep}{5pt}
\begin{tabular}{lccc}
\toprule
\textbf{Method} & \textbf{Interaction} & \textbf{Supervision} & \textbf{Primary use} \\
\midrule

Single-task RL & Single objective & Single reward & Nominal task learning \\ 
\textbf{RCRL} & \textbf{Single objective} & \textbf{Diverse rewards} & \textbf{Nominal task learning} \\ 
MTRL / MORL & Diverse objectives & Diverse rewards & Policy steerability \\ 
USFA & Diverse objectives & Successor features & Policy steerability \\ 
FB & Broad reward-free data & Occupancy representations & Policy steerability \\

\bottomrule
\end{tabular}

\label{tab:related_work}
\end{table*}

\textbf{Sample-efficient single-task RL.}
A body of work aims to improve the efficiency and stability of deep RL when learning a fixed objective, through techniques such as increased data reuse~\citep{d2022sample}, improved value estimation~\citep{bellemare2017distributional, fujimoto2018addressing}, exploration~\citep{ciosek2019better}, and scalable network architectures~\citep{nauman2024bigger, bhatt2024crossq}. Recent methods achieve strong performance by combining several of these advances~\citep{lee2025hyperspherical}. RCRL operates in the same setting and is complementary to these improvements, but rather than changing how transitions are collected or increasing the number of gradient steps, it extracts additional supervision from existing replay by evaluating each transition under diverse reward specifications. Our experiments therefore treat strong single-task algorithms as backbones and ask whether diverse reward supervision can further improve their learning efficiency.

\textbf{Multi-task and multi-objective RL.}
Multi-task RL (MTRL) trains shared agents across multiple tasks~\citep{teh2017distral,hessel2019multi,yu2020gradient,kumar2023offline,nauman2025bigger}, while multi-objective RL (MORL) learns policies across combinations of potentially competing objectives, often by conditioning on preferences or reward weights~\citep{roijers2013survey,yang2019generalized,abels2019dynamic,hayes2022practical}. RCRL uses related conditioning machinery but studies a different setting. Alternative reward specifications are used to relabel experience collected for a predefined nominal objective, rather than to define the objectives under which that experience must be collected. This separates reward diversity in supervision from reward diversity in exploration and lets us ask whether learning across objectives can improve the objective of interest itself.

\textbf{Goal-conditioned RL.} This approach defines tasks in terms of goal states, typically using sparse reward functions that indicate success with respect to a given goal~\citep{kaelbling1993learning, pmlr-v37-schaul15}. In techniques like HER, the agent is conditioned on a goal representation, allowing a single agent to learn behaviors for multiple goals within the same environment~\citep{andrychowicz2017hindsight, akkaya2019solving}, supporting adaptation to new goals~\citep{eysenbach2018diversity, eysenbach2022contrastive, park2024ogbench}. RCRL applies a closely related idea to reward specifications rather than goal states. RCRL stores the components required to evaluate a family of rewards, recomputes scalar rewards on replay, and optimizes them through the same objective as the nominal reward. Unlike HER, which is primarily motivated by learning from goal-reaching rewards, RCRL studies whether relabeling with alternative dense rewards can improve a known objective when exploring under a single objective.

\textbf{Successor features and reward-conditioned representations.}
Successor features (SFs) factor value functions into expected state-action features and reward weights, enabling transfer across rewards that are linear in a shared feature representation~\citep{barreto2017successor,barreto2018transfer}. Universal successor feature approximators (USFAs) extend this idea by conditioning successor features on task descriptors and using generalized policy improvement for transfer~\citep{borsa2018universal}. Forward-backward (FB) representations similarly learn reward-independent occupancy representations that can be mapped to new rewards after training~\citep{touati2021learning,touati2022does,zheng2026intention}. Additionally, recent approaches learn representations from broad offline data, often without rewards~\citep{frans2024unsupervised,ingebrand2024zero}. RCRL shares the goal of representing multiple reward-dependent behaviors, but uses a different learning mechanism and data regime. Firstly, it directly trains scalar reward-conditioned value functions from recomputed rewards. Furthermore, RCRL learns from data generated online by the known nominal objective rather than from broad reward-free or multi-objective exploration.

\section{Reward-Conditioned Reinforcement Learning}
\label{sec:method}

\begin{figure}[t]
  \begin{center}
\centerline{\includegraphics[width=0.95\linewidth]{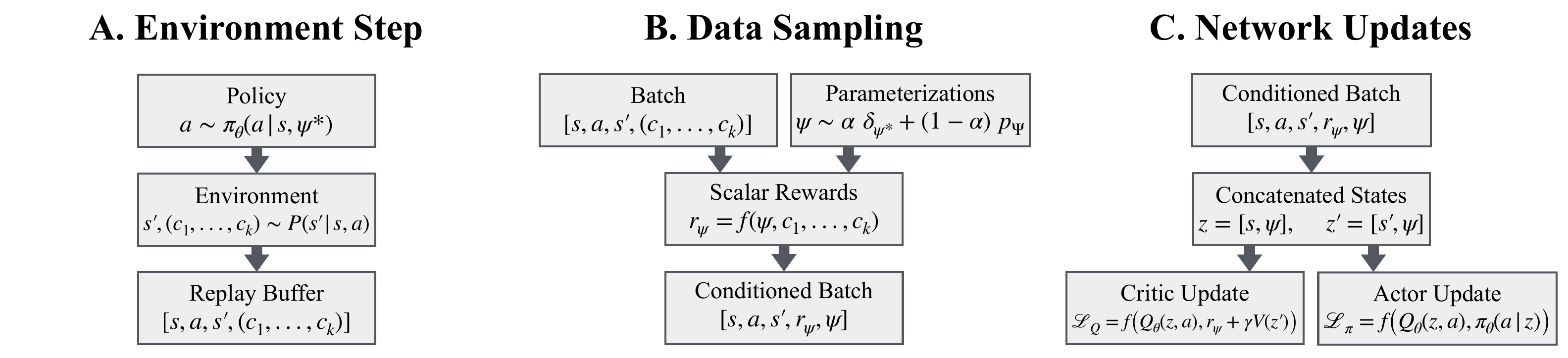}}
    \caption{\footnotesize \emph{\textbf{Overview of RCRL.}} During interaction, actions are sampled from the policy conditioned on the nominal reward $\psi^{\star}$, with reward components stored in the buffer. During training, each transition is paired with an independently sampled reward specification $\psi \sim \mathcal{P}_{\Psi}$ to compute the scalar reward $r_{\psi}$. The training then follows with the agent conditioned on $\psi$.}
    \label{fig:augmentation_scheme}
  \end{center}
\vspace{-0.1in}
\end{figure}

We consider rewards of the form $r_{\psi}(s,a)=f\bigl(\psi,c_1(s,a),\ldots,c_k(s,a)\bigr)$, where $c_1,\ldots,c_k$ are measurable reward components and $\psi$ specifies how they are combined into a scalar reward. Such structure is common in control, where rewards often combine terms for task progress, pose or velocity tracking, and control costs. Usually, single-task RL optimizes a fixed specification $\psi^\star$. Once the underlying components are available the same transition can be evaluated under many alternative specifications $\psi\in\Psi$. RCRL exploits this structure, as summarized in Figure~\ref{fig:augmentation_scheme} and Pseudocode~\ref{alg:reward_conditioned_rl}.

{\setlength{\textfloatsep}{5pt}
\begin{algorithm}[ht]
    \vspace{-0.03in}
    \begin{algorithmic}[1]
    \small
    \caption{Reward-Conditioned Reinforcement Learning}
    \label{alg:reward_conditioned_rl}
    
    \item \textbf{Input:} nominal objective $\psi^{\star}$, alternative specifications $\Psi$, mixture distribution $\mathcal{P}_{\Psi}$ 

    \vspace{-0.07in}
       \hrulefill
       
       \small \STATE $a \sim \pi(a|s,\psi^{\star})$ \quad \texttt{\# explore via a policy conditioned on $\psi^{\star}$}

       \vspace{0.01in}
       
       \small \STATE $s', c_1, ..., c_k = \textsc{env.step}(a)$ \quad \texttt{\# gather the reward components $c_1, ... c_k$ when acting}

       \vspace{0.01in}
       
       \small \STATE $\mathcal{D} \leftarrow \mathcal{D} \cup (s,a,c_1, ..., c_k,s')$ \quad \texttt{\# store rewards components in the buffer}

       \vspace{0.01in}
       
       \small \STATE $\{s ,a,c_1, ..., c_k,s'\}_{i} \sim \mathcal{D} ~~ \text{with} ~~ i = 1, ..., B$ \quad \texttt{\# sample B transitions from the buffer}
       
       \vspace{0.01in}
       
       \small \STATE $r_i = f(\psi_i, c_1, ..., c_k) ~~ \text{with} ~~ \psi_{i} \sim \mathcal{P}_{\Psi}$ \quad \texttt{\# evaluate sampled reward specifications}

       \vspace{0.01in}
       
       \small \STATE $\textsc{agent.update}(\{s,a,r_{\psi},s', \psi \}_{i})$ \quad \texttt{\# update the agent while conditioning on $\psi$}

    \end{algorithmic}
\end{algorithm}}

The nominal specification $\psi^\star$ defines both the target task and the reward used for environment interaction. RCRL additionally trains on alternative specifications $\psi\in\Psi$ by conditioning the policy and critic on the active specification, $\pi_\theta(a\mid s,\psi)$ and $Q_\phi(s,a,\psi)$. Transitions are stored together with their reward components, allowing their rewards to be recomputed under different specifications during replay. For each sampled transition, RCRL draws a reward specification according to:

\vspace{-0.12in} 
\[ \mathcal{P}_{\Psi} = \alpha~\delta_{\psi^{\star}} + (1 - \alpha)~p_{\Psi}, \] 
\vspace{-0.12in}

where $\delta_{\psi^\star}$ is a point mass on the nominal specification $\psi^\star$ and $p_{\Psi}$ samples alternative reward specifications $\psi$. The coefficient $\alpha\in[0,1]$ controls the balance between nominal and counterfactual supervision, with $\alpha=1$ recovering standard single-task RL. After sampling $\psi$, the corresponding reward is recomputed from the stored components and the transition is used to update the agent conditioned on $\psi$. We sample reward specifications independently for each transition in the batch. The alternative specifications affect learning but not the objective used to collect experience, and the agent always conditions on $\psi^\star$ (the nominal objective) for the environment interaction. RCRL therefore learns alternative reward-conditioned behaviors off-policy from nominal task experience. The primary question that we study \emph{is whether this counterfactual supervision improves learning of $\psi^\star$}. As we show in Section~\ref{sec:experiments}, it improves nominal performance across several settings, while the learned reward-conditioned policy also supports steering toward related reward specifications. 

\subsection{Constructing the Set of Reward Specifications}
\label{sec:constructing_the_set}

A central design choice in RCRL is how to construct the alternative reward specifications $\Psi$. We consider two complementary strategies that provide different forms of reward diversity.

\textbf{Perturbed Reward Conditioning (PRC).}
In the first approach, we generate continuous variations of the nominal reward specification $\psi^\star$. Many control rewards combine multiple components whose relative contributions are determined by reward parameters, naturally admitting variations. We exploit this structure by applying nonnegative elementwise perturbations $\Delta \in \mathbb{R}^k$ according to $\psi = \psi^\star \odot \Delta$. Sampling $\Delta$ from a distribution over nonnegative values centered at unity produces a family of rewards around the nominal objective while preserving the signs of its components. The spread of this distribution controls the diversity of reward supervision and nonnegative perturbations ensure that the qualitative structure of the reward is preserved, including component signs.

\textbf{Auxiliary Reward Conditioning (ARC).} In the second approach, we construct the set of alternative reward specifications $\Psi$ using reward functions corresponding to distinct tasks defined in the same state and action space. Unlike PRC, which generates variations of the nominal reward, ARC considers qualitatively different objectives applied to the same embodiment. This approach is natural in multi-task and multi-objective settings, where related tasks or objectives provide a readily available set of reward functions for relabeling. For example, when the nominal reward $\psi^\star$ defines a \texttt{run} objective, additional reward specifications in $\Psi$ may correspond to related behaviors such as \texttt{standing} or \texttt{walking}. As such, the auxiliary objectives might provide complementary signal or form learning curricula that facilitate the learning (e.g. standing facilitates walking, which in turn supports running).

\subsection{Reward Conditioning and Training Stability}

As illustrated in Figure~\ref{fig:augmentation_scheme}, RCRL conditions the agent on the reward specification used for each update. In this subsection, we discuss different representations used for this conditioning.

\textbf{Conditioning on the Reward Specification.}
For continuously parameterized rewards, such as PRC, the reward parameters $\psi$ or perturbation vector $\Delta$ can be concatenated directly to the environment state. For finite reward sets, such as ARC, we assign each reward specification a learnable embedding and provide this representation to both the actor and critic. The embeddings are learned jointly with the value function through the critic objective, following common practice in multi-task RL~\citep{nauman2025bigger}. In our main experiments, unless stated otherwise, we use embedding-based conditioning. We compare a variety of related conditioning approaches in Subsection~\ref{sec:what_makes_rcrl_effective}.

\textbf{Training Stability.}
Learning across reward specifications can be challenging when their reward scales or induced value functions differ substantially. We therefore retain the stabilization mechanisms of each backbone, including reward normalization, categorical value learning, and scaled critics in \textsc{SimbaV2} and \textsc{BRC}. Our ablations show that RCRL benefits particularly from these scaled critic designs (Figure~\ref{fig:scaling_results1}), indicating a strong interaction between reward conditioning and critic capacity. However, as our vision-based experiments show, \textsc{DrQV2} which uses a conventional scalar critic also benefits from RCRL. We provide further analysis in Subsection~\ref{sec:what_makes_rcrl_effective} and Appendix~\ref{app:training_stability}.

\section{Experiments}
\label{sec:experiments}

Our experiments ask whether diverse reward supervision improves learning of a fixed nominal objective. We first measure nominal task gains across benchmarks, then study key RCRL design choices and reward compatibility, and finally examine policy steerability beyond the nominal objective. All presented figures show $95$\% confidence intervals calculated using bootstrapping according to the RLiable methodology~\citep{agarwal2021precipice}. Additional experimental settings are provided in Appendix~\ref{app:experimental_details}.

\textbf{RCRL Setup.} In our experiments, the set of alternative reward specifications $\Psi$ is constructed using one of two strategies introduced in Section~\ref{sec:constructing_the_set}, PRC or ARC. Unless otherwise stated, we set $\alpha=0.5$, so that approximately half of the training updates use the nominal reward specification $\psi^\star$ and half use alternative specifications. For PRC, $\Psi$ is generated by multiplicatively perturbing the coefficients of the nominal reward. The perturbations $\Delta_i$ are sampled from a stratified log-uniform distribution with support $[0.25,4.0]$. For ARC, $\Psi$ is constructed from task-specific reward functions that can be evaluated on the same state-action transitions as the nominal task. We discuss other implementation details, including exact counterfactual reward definitions for both ARC and PRC in Appendix~\ref{app:implementation_details}.

\subsection{RCRL Improves Learning Efficiency}
\label{sec:nominal_performance}

We first evaluate whether diverse reward supervision improves learning of the nominal objective. Across all experiments in this section, we follow the training protocols and hyperparameters of the corresponding backbone and evaluate performance exclusively under the nominal reward. We summarize these runs in Figure~\ref{fig:iclr_timeseries} and present detailed training curves in Appendix~\ref{app:training_curves}.

\begin{figure}[ht]
  \centering
  
  \begin{subfigure}{\linewidth}
    \centering
    \includegraphics[width=0.99\linewidth]{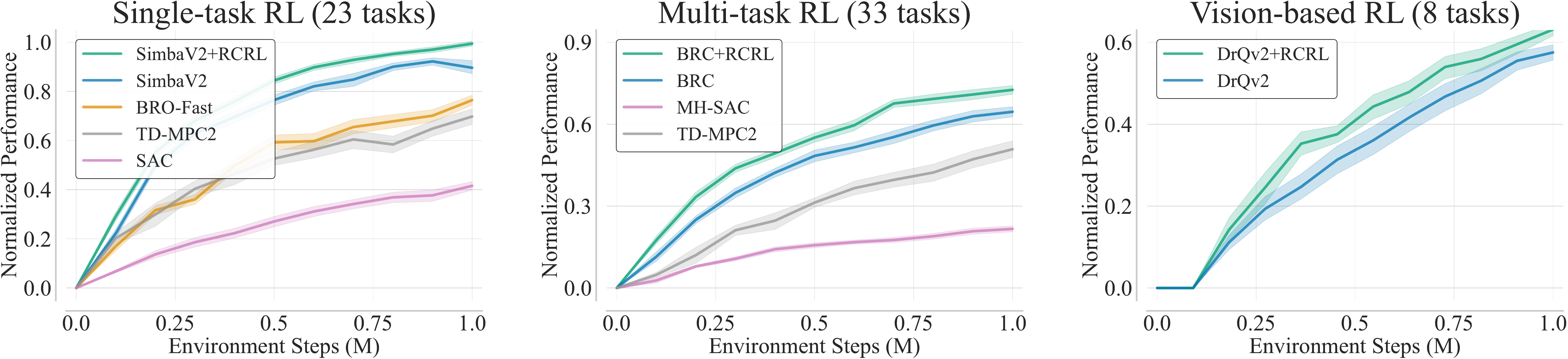}
  \end{subfigure}
  \vspace{-0.1in}
  \caption{\footnotesize \emph{\textbf{RCRL improves learning efficiency.}} The figure presents aggregate training performance across single-task, multi-task, and vision-based control. Adding counterfactual reward supervision via RCRL improves the corresponding backbone while evaluating under the nominal reward. We provide additional experimental details in Appendix~\ref{app:experimental_details}.}
  \label{fig:iclr_timeseries}
\end{figure}

\textbf{Single-task RL.} We first augment \textsc{SimbaV2}~\citep{lee2025hyperspherical} with PRC and evaluate it on the $23$-task continuous-control tasks used in the original study, spanning DMC~\citep{tassa2018deepmind}, HumanoidBench~\citep{sferrazza2024humanoidbench}, and OpenAI Gym~\citep{openaigym}, and using $10$ seeds. We list all tasks in Appendix~\ref{app:benchmarks}. We preserve the original \textsc{SimbaV2} training setup, including the training horizon and hyperparameters, and introduce only the reward-conditioning components required by RCRL. As shown in the left panel of Figure~\ref{fig:iclr_timeseries}, RCRL improves both final and AUC performance by around $10\%$. These results show that diverse reward supervision via RCRL can improve learning of a fixed nominal objective even when both data collection and evaluation remain focused on the nominal objective.

\begin{wrapfigure}[13]{r}{0.48\textwidth}
  \centering
  \vspace{-9pt}
  \includegraphics[width=0.425\textwidth]{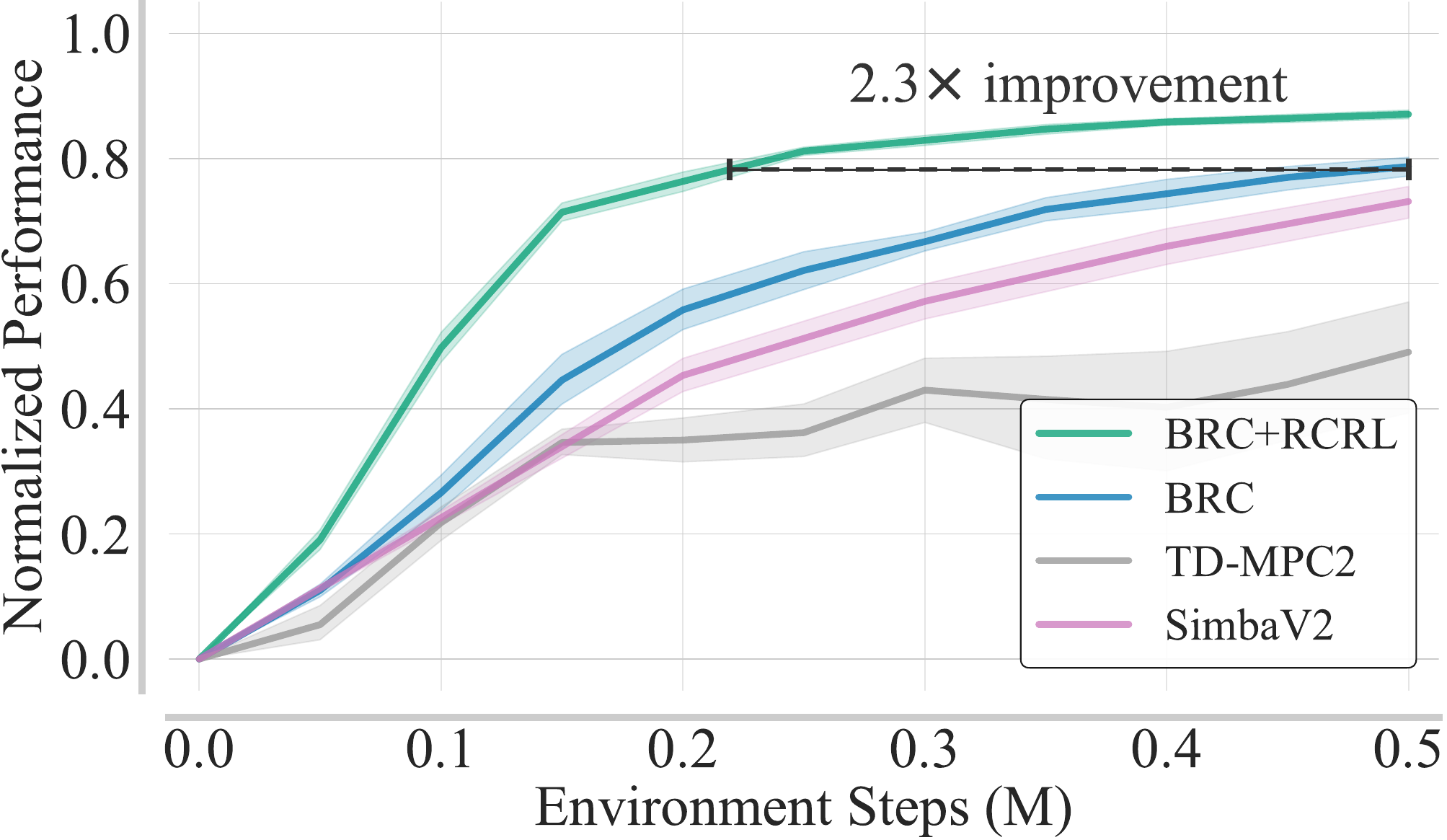}
  \vspace{-5pt}
  \caption{\footnotesize \emph{\textbf{Learning on DMC Dogs.}} RCRL yields a $2.3\times$ improvement in sample efficiency, reaching $80\%$ performance after approximately $200$k environment steps.}
  \label{fig:dogs_results}
\end{wrapfigure}

\textbf{Multi-task RL.} We next augment \textsc{BRC}~\citep{nauman2025bigger} with ARC and evaluate it on DMC Dogs, HumanoidBench, and HumanoidBench Hard following the original work. We list all tasks in Appendix~\ref{app:benchmarks} and use $5$ seeds. We preserve the original \textsc{BRC} training protocol and hyperparameters, and introduce only the components required by RCRL. In this setting, each task continues to collect experience under its own nominal objective, while RCRL additionally evaluates the reward functions of related tasks on replayed transitions. As shown in the middle panel of Figure~\ref{fig:iclr_timeseries}, RCRL improves learning, showing that counterfactual supervision remains beneficial even when applied on top of a strong multi-task learner. ARC is attractive in this setting because trajectories collected for one task can also be evaluated under the reward functions of the other tasks. For $N$ tasks, this creates up to $N(N-1)$ additional pairings from the same environment interactions. As shown on Figure~\ref{fig:dogs_results}, the effect is pronounced on the DMC Dogs. There, \textsc{BRC+RCRL} learns substantially faster than the original multi-task learner, leading to more than twofold improvement in sample efficiency and reducing the variance across seeds. This result shows that diverse reward supervision via RCRL can complement the benefits of multi-task training.

\textbf{Vision-based RL.} We also apply PRC to \textsc{DrQV2}~\citep{yarats2021drqv2} on $9$ image-based DMC tasks listed in Appendix~\ref{app:benchmarks}. As shown in the right panel of Figure~\ref{fig:iclr_timeseries}, RCRL improves final performance and AUC by $10\%$ and $17\%$ respectively. We find that RCRL improves learning performance in this setup as well, showing that the effect is not limited to proprioceptive observations or to the architectural choices used by \textsc{SimbaV2} or \textsc{BRC}. This provides complementary evidence that counterfactual reward supervision remains beneficial across observation modalities and learning architectures.

\begin{tcolorbox}[takeawaybox]
  Broad, diverse reward supervision via RCRL can outperform specialized training even when the evaluation is restricted to a single predefined nominal objective.
\end{tcolorbox}

\subsection{What Makes RCRL Effective?}
\label{sec:what_makes_rcrl_effective}

Previous subsection shows that diverse reward supervision via RCRL improves nominal task learning. In this subsection, we investigate design choices that drive these improvements. We first disentangle the benefits of diverse reward supervision from broader state-action coverage, then investigate the roles of reward conditioning, exploration, sampling balance, and model capacity.

\begin{figure}[ht]
  \centering
  
  \begin{subfigure}{0.49\linewidth}
    \centering
    \includegraphics[width=0.99\linewidth]{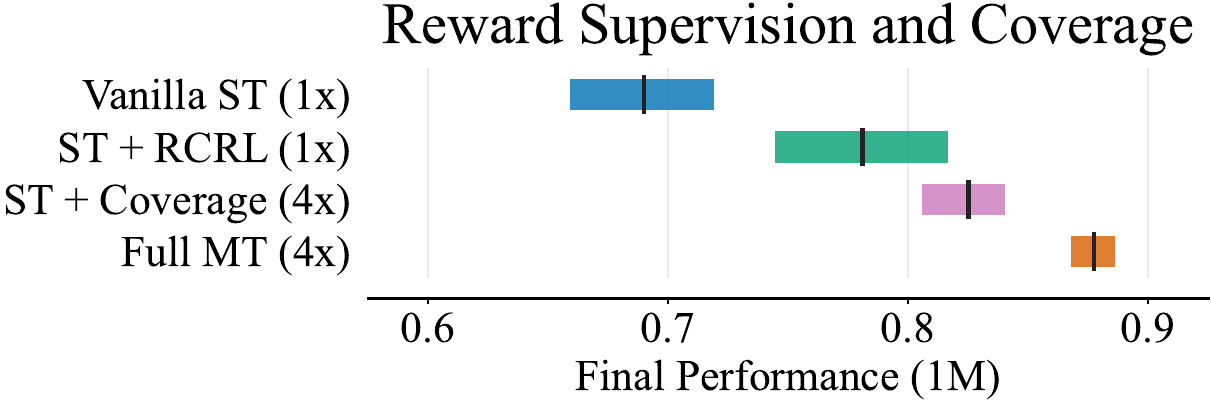}
    \end{subfigure}
    \hfill
  \begin{subfigure}{0.49\linewidth}
    \centering
    \includegraphics[width=0.99\linewidth]{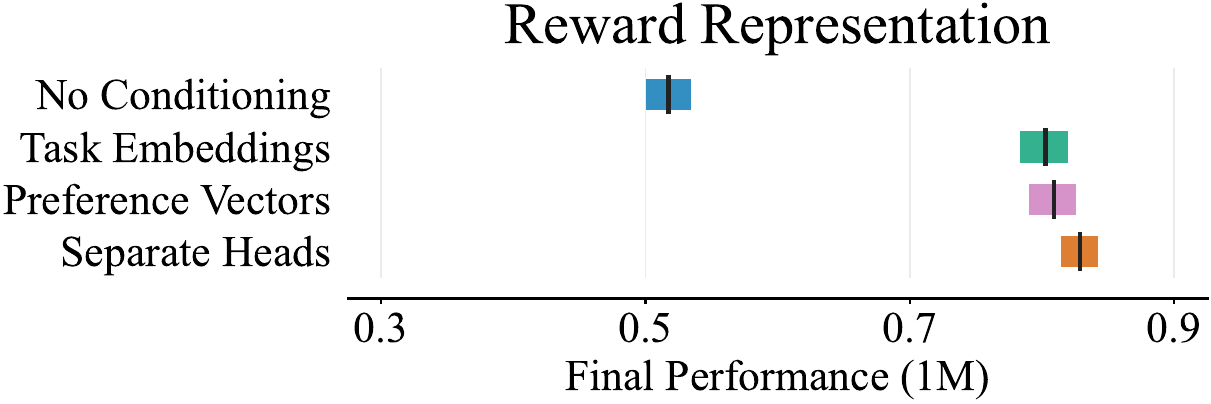}
    \end{subfigure}
\vspace{-0.05in}
\caption{\footnotesize
\emph{\textbf{Ablation studies.}}
The left panel shows that improvements from multi-task learning come from both broader state-action coverage and additional reward supervision; RCRL captures the latter without additional environment interaction.
The right panel shows ablations on conditioning the agent on $\Psi$. Removing conditioning degrades performance significantly, however, all tested methods that condition the agent on the reward information perform similarly well.}
  \label{fig:results_ablations1}
\end{figure}

\textbf{Reward supervision and coverage.}
We next disentangle two sources of multi-task gains: broader state-action coverage and diverse reward supervision. We test \textsc{BRC} on the four DMC Dogs tasks, training for $1$M environment steps per task with $5$ random seeds. We compare single-task training (Vanilla ST), single-task RCRL (ST+RCRL), single-task training with expanded state-action coverage (ST+Coverage), and full multi-task training (Full MT). Vanilla ST and ST+RCRL use only experience collected under the nominal task and are matched in interaction. ST+Coverage additionally receives replay from single-task runs on the other three tasks, giving it access to $4\times$ as many transitions while still optimizing only the nominal reward. Following tandem RL~\citep{ostrovski2021difficulty}, these transitions are progressively added during training. Full MT uses the same broader experience while also learning from the reward associated with each task. As shown in the left panel of Figure~\ref{fig:results_ablations1}, both ST+RCRL and ST+Coverage outperform Vanilla ST. Full MT, which combines broader coverage and diverse reward supervision performs best. This shows that both factors contribute to multi-task gains, while RCRL captures the benefit of diverse reward supervision from the original interaction budget.

\textbf{Reward representation.}
RCRL requires the agent to distinguish which reward specification is active, but this conditioning can be implemented in several ways. We compare no conditioning, fixed preference vectors, learned reward embeddings following \textsc{BRC} recipe~\citep{nauman2025bigger}, and separate reward-specific heads, while keeping the rewards and training data fixed under ARC variant of \textsc{SimbaV2+RCRL}. We evaluate all variants across the $18$ tasks from DMC Dogs and HumanoidBench for $1$M steps using $10$ seeds. As shown in the right panel of Figure~\ref{fig:results_ablations1}, removing conditioning degrades performance, whereas the three conditioned variants perform similarly. This indicates that identifying the active reward is essential for benefiting from diverse supervision, while the particular representation of that identity is comparatively unimportant. In practice, RCRL therefore does not depend on a specialized conditioning architecture.

\begin{wrapfigure}[13]{r}{0.48\textwidth}
  \centering
  \vspace{-10pt}
  \includegraphics[width=0.425\textwidth]{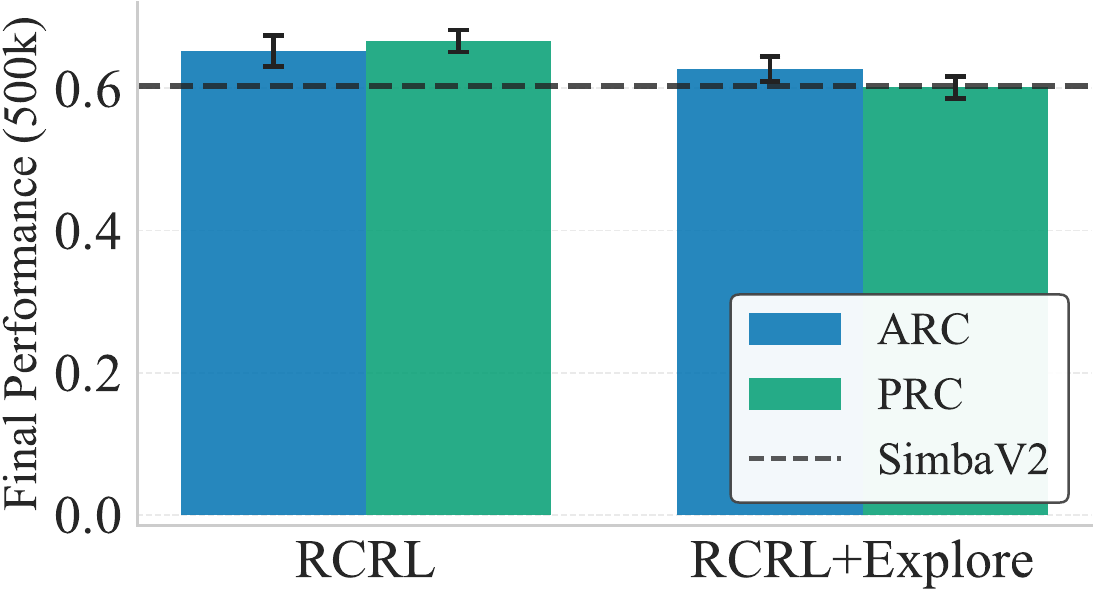}
  \vspace{-3pt}
  \caption{\footnotesize \emph{\textbf{RCRL and exploration.}} RCRL performs best when exploration remains nominal. RCRL+Explore performs worse than RCRL but outperforms vanilla \textsc{SimbaV2}.}
  \label{fig:sup_vs_explore}
\end{wrapfigure}

\textbf{Reward supervision and exploration.}
We next test whether alternative rewards should also influence environment interaction. Across the same $18$ tasks, we compare single-task \textsc{SimbaV2}, \textsc{SimbaV2+RCRL}, and \textsc{SimbaV2+RCRL+Explore}, evaluating both ARC and PRC approaches, and training each for $500$k environment steps with $10$ random seeds. Standard RCRL collects experience only under the nominal reward, whereas RCRL+Explore samples the reward function used for exploration from the same distribution used for training, and resampling every environment step. We evaluate both ARC and PRC while keeping other settings fixed, isolating the effect of reward-diverse exploration. As shown in Figure~\ref{fig:sup_vs_explore}, Standard RCRL achieves the highest nominal performance for both ARC and PRC. At the same interaction budget, RCRL+Explore also slightly outperforms \textsc{SimbaV2}. Thus, reward diversity can be beneficial even when it affects exploration, but using diverse reward supervision paired with task-focused exploration yields the strongest nominal performance.

\textbf{Sensitivity to counterfactual supervision.} We next investigate how the balance between nominal and counterfactual updates affects RCRL. The mixture coefficient $\alpha$ controls the probability of sampling the nominal reward for each update, allowing us to vary the amount of counterfactual supervision. We evaluate $\alpha \in \{0.1,0.3,0.5,0.7,0.9\}$, with $\alpha=0.5$ used by default, using \textsc{SimbaV2} backbone on the same $18$ tasks, $500$k steps and $10$ seeds. As shown in Figure~\ref{fig:results_ablation3}, RCRL improves performance across a broad range of sampling ratios, with the strongest results at $\alpha=0.7$, corresponding to $70\%$ nominal and $30\%$ samples relabeled with counterfactual rewards. This suggests that counterfactual supervision remains beneficial even when it replaces a substantial fraction of nominal updates, although emphasizing the nominal objective yields the best performance. We further examine the choice and spread of the alternative reward sampling distribution in Appendix~\ref{app:experimental_setup_ablation}, finding that RCRL remains effective across various distributions.

\begin{wrapfigure}[13]{r}{0.48\textwidth}
  \centering
  \vspace{-10pt}
  \includegraphics[width=0.425\textwidth]{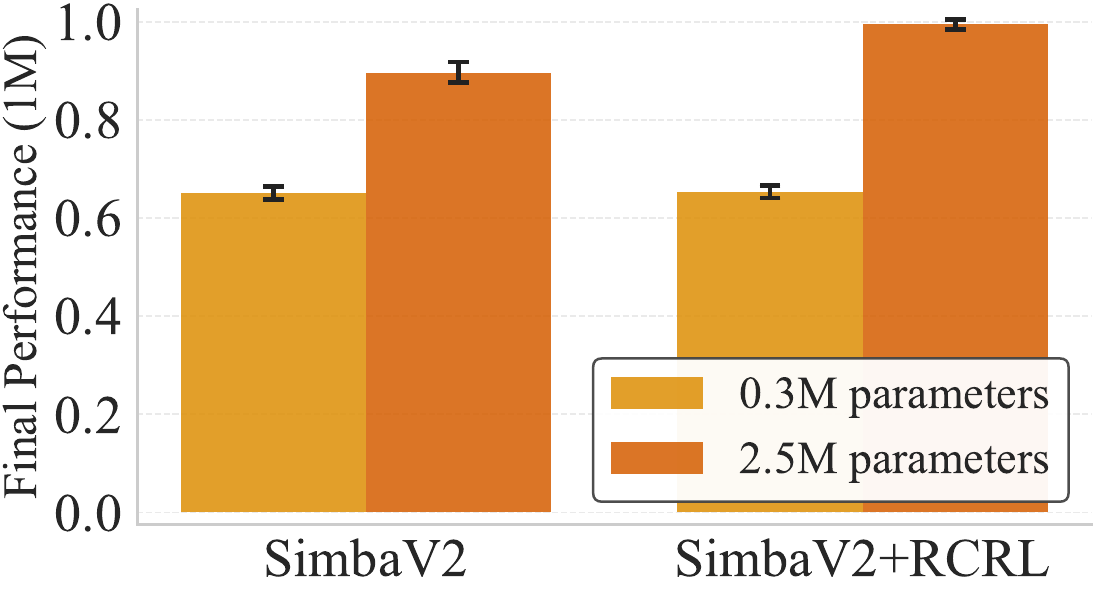}
  \vspace{-2pt}
  \caption{\footnotesize \emph{\textbf{RCRL and model capacity.}} Diverse supervision improves performance when the critic has sufficient capacity to model counterfactual rewards.}
  \label{fig:scaling_results1}
\end{wrapfigure}

\textbf{Model capacity.}
Finally, we study how RCRL interacts with critic capacity. Diverse reward supervision requires a critic to represent value functions for many specifications, suggesting that it may only be useful when the model has sufficient capacity. Figure~\ref{fig:scaling_results1} compares \textsc{SimbaV2} and \textsc{SimbaV2+RCRL} across 23 tasks considered in the single-task benchmark (Appendix~\ref{app:benchmarks}), trained for $1$M steps, using widths of $128$ ($0.3$M parameters) and $512$ ($2.5$M parameters), with the latter corresponding to the default. At smaller scale, RCRL provides no improvement over the baseline. In contrast, RCRL improves performance significantly when paired with a scaled model. Interestingly, increased capacity benefits both methods, but the gain is larger for RCRL. This suggests that scale enables the agent to exploit the richer supervision provided by counterfactual rewards. Thus, RCRL and model capacity appear to be complementary, with RCRL being the most effective once the learner has sufficient capacity to represent it.

\begin{tcolorbox}[takeawaybox]
    RCRL achieves the strongest nominal task performance with task-focused exploration, although reward-diverse exploration can still outperform standard single-task RL. Explicit reward conditioning is essential, and sufficient model capacity helps exploit the benefits of diverse supervision.
\end{tcolorbox}

\subsection{RCRL and Reward Compatibility}
\label{sec:reward_compatibility}

Although RCRL improves nominal task learning, not all alternative rewards have to be equally useful. This raises the question of what makes a counterfactual reward beneficial for a given task. In this subsection, we investigate this question by measuring the contributions of various reward functions.

\begin{figure}[t]
  \centering
  
  \begin{subfigure}{\linewidth}
    \centering
    \includegraphics[width=1.0\linewidth]{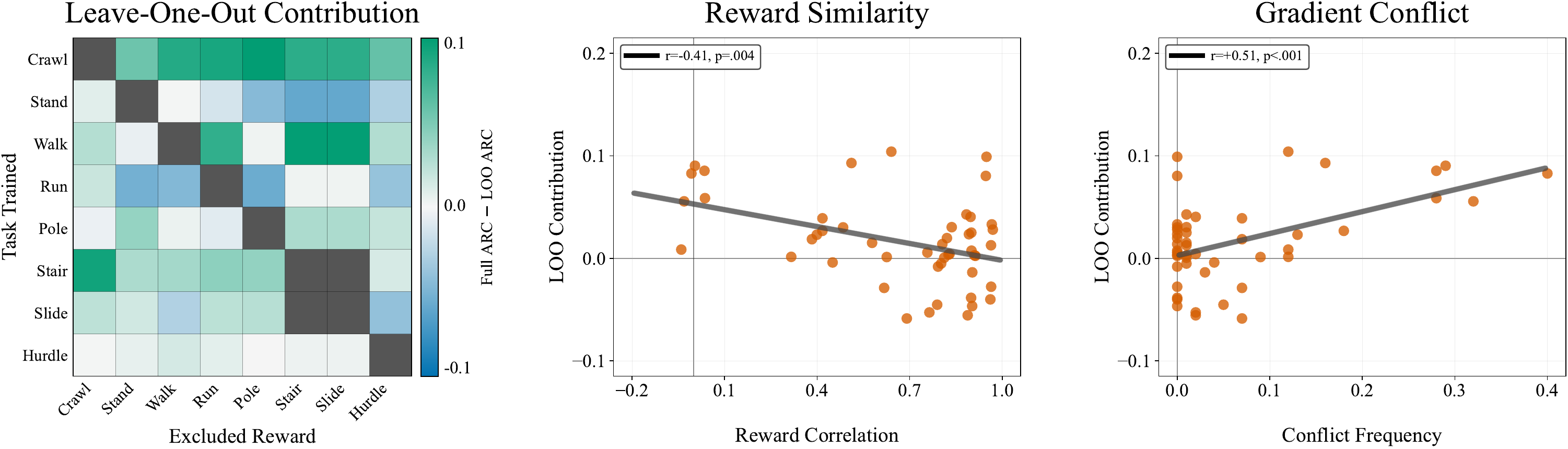}
  \end{subfigure}
  \vspace{-0.15in}
  \caption{\footnotesize
  \emph{\textbf{RCRL and reward compatibility.}}
  The left panel shows the leave-one-out contribution $C_{i,j}$ of each counterfactual reward to nominal task learning. The middle and right panels relate these contributions to reward correlation and gradient conflict frequency respectively. In both cases, less similar signals tend to make larger contributions in RCRL.}
  \label{fig:iclr_heatmap}
\end{figure}

\textbf{Leave-one-out reward analysis.}
We first measure the contribution of individual alternative rewards across the $8$ HumanoidBench tasks listed in Table~\ref{tab:auxiliary_tasks}, using $1$M steps and $10$ seeds. For each nominal task, we compare \textsc{SimbaV2+RCRL} trained with the full ARC reward set against variants in which one alternative reward is removed. We keep all other aspects of training fixed. Since \texttt{stair} and \texttt{slide} tasks share the same reward function, the $8$ tasks correspond to seven distinct alternative reward specifications. We define the contribution of reward $j$ to nominal task $i$ as $C_{i,j}=\mathrm{AUC}_{i}-\mathrm{AUC}_{i\setminus j}$, where positive values indicate that including reward $j$ improves nominal task learning and negative values indicate interference. As shown in the left panel of Figure~\ref{fig:iclr_heatmap}, the mean contribution across task-alternative reward pairs is positive, corresponding to a $2\%$ improvement, but individual effects vary substantially from $-5\%$ to $10\%$. Compatibility is task dependent, with \texttt{crawl} benefiting from every reward considered, and some rewards reducing performance on \texttt{stand} or \texttt{run}. Thus, the benefit of diverse reward supervision depends on the nominal task and the choice of counterfactual rewards.

\textbf{Reward similarity.} We next ask whether the most useful counterfactual rewards are simply those that most closely resemble the nominal reward. For all $48$ task-alternative reward pairs, we measure the correlation $R_{i,j}$ between the nominal and counterfactual reward signals evaluated on the same transitions. We aggregate both reward correlation and leave-one-out contribution over the training. As shown in the middle panel of Figure~\ref{fig:iclr_heatmap}, the opposite pattern emerges, with low correlation rewards making larger contributions, with a Pearson correlation of $r=-0.41$. This suggests that useful counterfactual rewards need not reproduce the nominal reward, and may improve learning the nominal objective by providing distinct supervision that is qualitatively different from the nominal reward.

\textbf{Gradient compatibility.} Counterfactual rewards may be useful precisely when they induce optimization signals that differ from the nominal objective. In particular, gradient conflict may indicate that an alternative reward provides complementary rather than redundant supervision. To examine this, for each of the $48$ task-alternative reward pairs considered in the previous paragraph we sample the same transitions, compute critic gradients under the nominal and counterfactual rewards, and measure the frequency of gradient conflict with respect to the critic parameters calculated according to \citet{yu2020gradient}. Specifically, we calculate the dot product between nominal and counterfactual reward minibatches gradients in the critic backbone, and we denote conflict when this dot product is negative. Similarly to the previous paragraph, we aggregate conflict frequency and leave-one-out contribution over training for the same task-alternative reward pairs. As shown in the right panel of Figure~\ref{fig:iclr_heatmap}, alternative rewards that induce gradient conflict more frequently tend to make larger contributions to nominal task learning. This relationship is strong with a correlation of $r=0.51$. Interestingly, this contrasts with the intuition that useful auxiliary objectives should produce gradients aligned with the nominal objective. Instead, our results suggest that diverse and even conflicting signals can provide complementary supervision and improvements.

\begin{wrapfigure}[13]{r}{0.48\textwidth}
  \centering
  \vspace{-10pt}
  \includegraphics[width=0.425\textwidth]{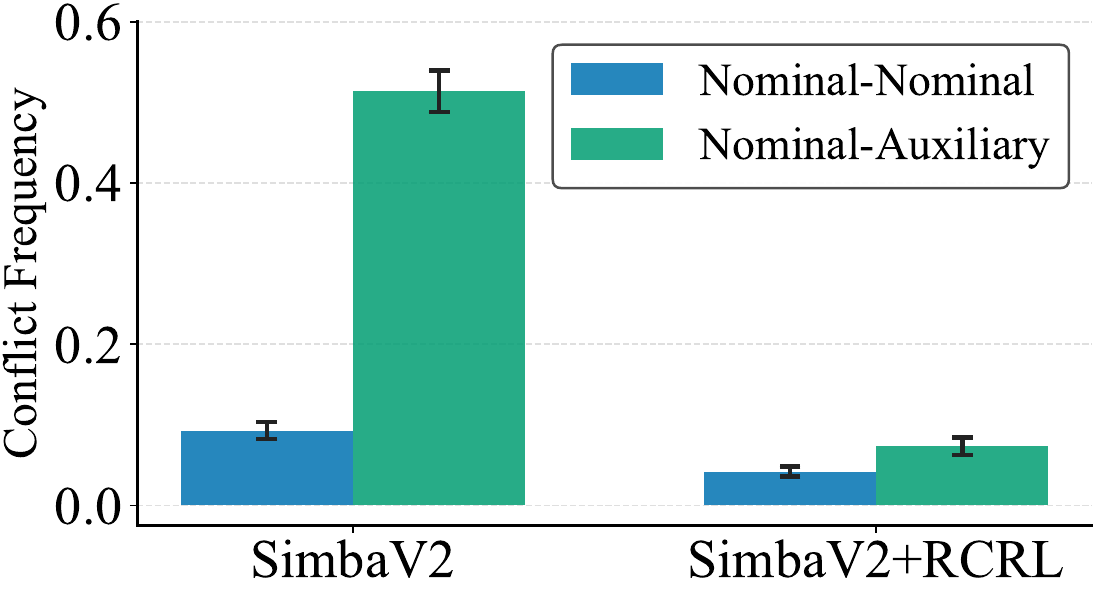}
  \vspace{-3pt}
  \caption{\footnotesize \emph{\textbf{RCRL and gradient consistency.}} RCRL reduces nominal-counterfactual conflict, but also reduces nominal-nominal conflict between independently sampled batches.}
  \label{fig:gradient_conflict_explanation}
\end{wrapfigure}

\textbf{RCRL improves gradient consistency.}
The previous analysis shows that beneficial counterfactual rewards often induce gradient conflicts with the nominal objective. However, this raises a question: \textit{does training with potentially conflicting rewards lead to a critic whose gradients become more consistent over time?} To investigate this, we compare gradient conflict under single-task and RCRL training across the same eight tasks, considering both nominal–counterfactual gradients and gradients computed from independent nominal-reward minibatches (Figure~\ref{fig:gradient_conflict_explanation}). We find that nominal–counterfactual conflict decreases from $0.51$ to $0.07$ under RCRL. Interestingly, conflict between independent nominal-reward minibatches also decreases, from $0.09$ to $0.04$. These results suggest that learning from diverse reward specifications is associated with more consistent critic updates, even when evaluated exclusively on the nominal objective. One possible explanation is that counterfactual supervision encourages representations that accommodate multiple reward signals while improving gradient alignment within the nominal task. Although this analysis does not establish a causal mechanism, it provides evidence that the benefits of diverse reward supervision are accompanied by changes in the critic's optimization dynamics.

\begin{tcolorbox}[takeawaybox]
    Useful alternative rewards need not resemble the nominal objective. Less correlated rewards and more conflicting gradients are associated with larger nominal task improvements. Despite this conflicting supervision, RCRL ultimately reduces gradient conflict even in nominal task batches.
\end{tcolorbox}

\subsection{Steerability Beyond the Nominal Objective}
\label{sec:steerability}

The main goal of RCRL is to improve a fixed predefined objective, but reward conditioning paired with diverse reward supervision also makes the policy steerable. In this subsection, we investigate whether behaviors associated with $\Psi$ can be recovered without further training.

\begin{figure}[ht]
    
  \begin{subfigure}{\linewidth}
    \centering
    \includegraphics[width=0.95\linewidth]{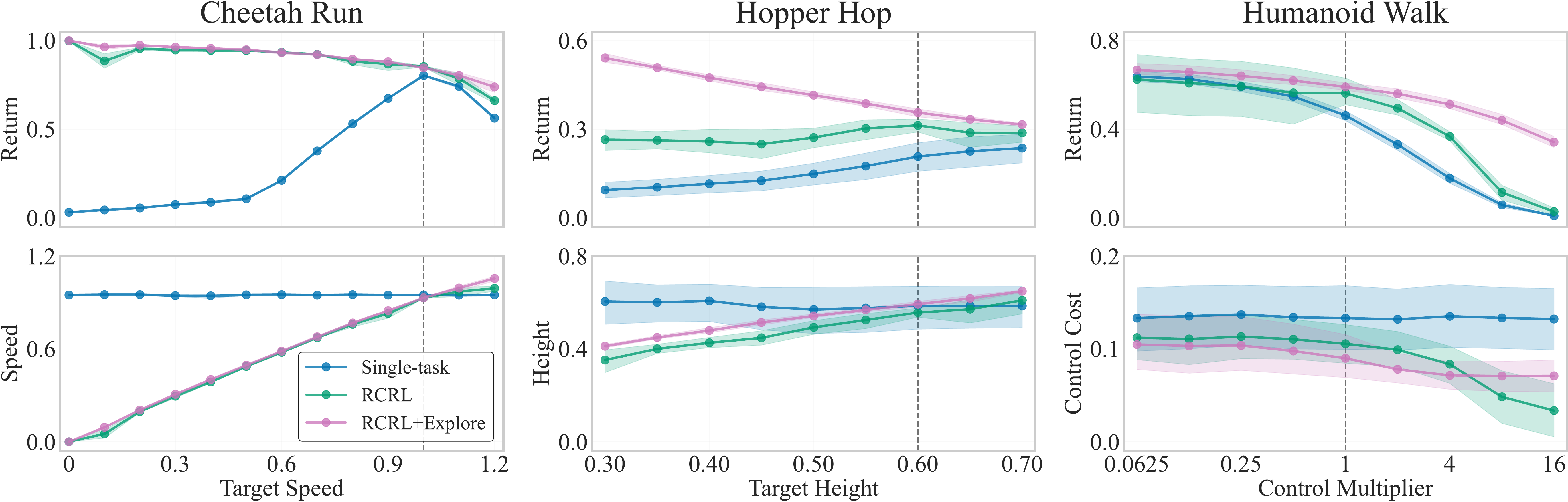}
    \end{subfigure}
    \vspace{-0.2in}
    \caption{\footnotesize
    \emph{\textbf{Zero-shot steerability.}}
    Alternative reward specifications induce different behaviors in \texttt{cheetah-run}, \texttt{hopper-hop}, and \texttt{humanoid-walk} tasks. We compare single-task \textsc{SimbaV2}, \textsc{RCRL}, and RCRL+Explore, which trains and explores across multiple rewards, as well as uses $3 \times$ more environment interactions. The dashed line denotes the nominal reward specification which is used to generate the experience for single-task and \textsc{RCRL}. \textsc{RCRL} produces steerable behavior, despite learning the other objectives entirely off-policy from nominal task experience.}
  \label{fig:results_zeroshot}
\end{figure}

\textbf{Zero-shot steerability.}
We evaluate RCRL on $3$ DM Control tasks: \texttt{cheetah-run}, with rewards targeting different running speeds; \texttt{hopper-hop}, with different standing heights; and \texttt{humanoid-walk}, with different action penalties. We compare \textsc{SimbaV2}, \textsc{SimbaV2+RCRL}, and \textsc{SimbaV2+RCRL+Explore}. Single-task and RCRL agents use $1$M environment steps, whereas RCRL+Explore uses $3$M steps, with $10$ seeds per domain. RCRL interacts only under the nominal objective and learns alternative rewards through relabeling, while RCRL+Explore corresponds to a full multi-objective training with expanded exploration under the alternative objectives. At evaluation, we vary only the reward conditioning, without further policy updates. As shown in Figure~\ref{fig:results_zeroshot}, RCRL achieves the highest average nominal performance while providing greater zero-shot steerability than single-task training. On alternative objectives, average performance increases from $0.26$ for single-task to $0.58$ with RCRL, approaching the performance of full multi-objective training with expanded state-action coverage. Consistent with off-policy learning, the steerability decreases as the requested objective moves further from the nominal task. As such, RCRL can recover behaviors when they are sufficiently supported by the nominal objective, whereas multi-task training directly explores conditioned on alternative objectives.

\textbf{Multi-objective and reward-free methods.}
We next compare RCRL with methods designed for learning across multiple objectives, including Preference Conditioned Control implemented as a variant of MORL with relabeling~\citep{yang2019generalized}, USFA~\citep{borsa2018universal}, and FB~\citep{touati2022does}. Since FB usually assumes access to an offline dataset, we initialize it with replay data collected during an RCRL training run. Using \textsc{SimbaV2} as the backbone, we evaluate all methods on all \texttt{walker}, \texttt{dog}, and \texttt{humanoid} tasks from DMC, leading to $10$ tasks in total, under a budget of $500$k environment interactions, reporting AUC on both nominal and alternative objectives from $5$ random seeds. RCRL achieves the strongest nominal task performance, improving over single-task \textsc{SimbaV2} while substantially increasing alternative task performance. PCC with counterfactual reward relabeling achieves slightly better alternative performance, but lower nominal performance, illustrating the trade-off between broad objective coverage and task-focused exploration. USFA and FB perform substantially worse in this low-budget setting, although FB is typically trained with much larger reward-free datasets.  Overall, these results show that reward-diverse exploration favors broader alternative task coverage, whereas nominally directed exploration better optimizes the predefined objective while retaining useful steerability.

\textbf{Transfer to alternative objectives.} Finally, we investigate whether alternative reward supervision also improves transfer adaptation to these alternative tasks. While improving nominal task  remains the primary goal of RCRL, exposure to alternative rewards may facilitate learning representations aligned with these rewards. Using the HumanoidBench tasks listed in Table~\ref{tab:auxiliary_tasks}, we train each agent for $1$M steps on a nominal source task and then finetune it for $500$k steps on each remaining target task, using $10$ seeds per source-target pair. During source-task training, RCRL receives counterfactual supervision from the target rewards, whereas the baseline learns only from the nominal reward. We measure the transfer benefit as
$T_{i,j}=\mathrm{AUC}^{RCRL}_{i\rightarrow j}-\mathrm{AUC}^{Base}_{i\rightarrow j}$.
As shown in Figure~\ref{fig:transfer_heatmaps2}, the individual effects vary across source-target pairs, ranging from $-12\%$ to $25\%$ improvements. Interestingly, we observe substantial variation in transfer effectiveness across source-target task pairs, with some tasks serving as more effective transfer sources than others. In our setting, transfer difficulty is further influenced by differences in environment layout across tasks (e.g. empty scene in the walk task versus obstacles in pole task), since policies only receive information on the angular velocities and reward specification, and do not receive explicit information about environment variations. These results suggest that diverse reward supervision via RCRL can facilitate downstream adaptation, but understanding when and why these benefits emerge requires further investigation.

\begin{tcolorbox}[takeawaybox]
    Diverse reward supervision via RCRL improves nominal task performance while enabling meaningful policy steerability. Exploring under alternative objectives can further improve auxiliary-task performance, but comes at the cost of nominal task efficiency.
\end{tcolorbox}

\section{Conclusions}
\label{sec:conclusions}

We introduced RCRL, an off-policy method for training agents on multiple reward specifications while collecting experience under a nominal objective. The central idea is simple: when rewards are structured, replayed transitions can be reused to compute counterfactual rewards, providing additional supervision beyond the nominal reward. By conditioning the agent on the reward specification used in each update, RCRL allows this supervision to improve learning without collapsing distinct objectives into a single policy. Across various benchmarks, RCRL improves learning under the nominal reward and supports policy steerability at deployment. These results show that reward conditioning is useful not only for representing multiple objectives, but also as a practical mechanism for improving single-task learning. We discuss the limitations of our work in Appendix~\ref{sec:limitations}

\section*{Acknowledgments}

We gratefully acknowledge the Polish high-performance computing infrastructure, PLGrid (HPC Center: ACK Cyfronet AGH), for providing computational resources and support under grant no. PLG/2025/018597. We would also like to thank the Python~\citep{van1995python}, NumPy~\citep{harris2020array}, Matplotlib~\citep{hunter2007matplotlib}, SciPy~\citep{virtanen2020scipy} and JAX~\citep{bradbury2018jax} communities for developing tools that supported this work. Marek Cygan was partially supported by National Science Centre, Poland, under the grant 2024/54/E/ST6/00388. Pieter Abbeel holds concurrent appointments as a Professor at UC Berkeley and as an Amazon Scholar. This paper describes work performed at UC Berkeley and is not associated with Amazon. 

\bibliography{bibliography}
\bibliographystyle{neurips_2026}

\newpage
\appendix

\subsection*{Impact Statement}

This work contributes to improving the robustness, adaptability, and efficiency of reinforcement learning systems by enabling agents to learn and adapt to multiple reward specifications without additional environment interaction. We do not foresee any direct societal impacts specific to this work beyond those generally associated with advances in reinforcement learning research.

\section{Limitations}
\label{sec:limitations}

RCRL requires alternative rewards to be evaluable on replayed transitions and is limited by the behavioral coverage of nominal task data. We focus on structured rewards with explicit components, but the same idea could extend to learned reward families. Objectives requiring states or actions rarely visited by the nominal policy may be difficult to learn without additional exploration. As shown in our experiments, the choice of alternative rewards also matters, with some providing useful supervision, and others interfering with learning. Although RCRL is robust across several perturbation distributions and scales, selecting informative and compatible reward specifications remains an open problem. Finally, richer reward supervision may require sufficient model capacity. Our ablations show larger models benefit more from RCRL. In our experiments, the added computational cost is modest: updates are approximately $10\%$ slower on fixed A$100$ hardware, while replay memory increases by only $O(k)$ float32 values per transition for $k$ reward components.

\section{Background}
\label{sec:background}

The goal of RL is to learn a policy that maximizes expected return in a MDP~\citep{puterman1994markov} $(\mathcal{S}, \mathcal{A}, P, r, \rho, \gamma)$. $\mathcal{S}$ and $\mathcal{A}$ denote the state and action spaces, $P(s' | s,a)$ the transition dynamics, $\rho$ the initial state distribution, $r(s,a)$ the scalar reward, and $\gamma \in (0,1)$ the discount factor. A policy $\pi(a | s)$ is optimized to maximize expected discounted return. The value $V^\pi(s)$ and Q-value functions $Q^\pi(s,a)$ denote the expected return from a state and state-action pair. In many control tasks, desired behavior is specified through composite reward functions~\citep{tassa2018deepmind, peng2018deepmimic}, which combine multiple components such as task progress, control cost or smoothness. In such case, the scalar reward $r(s,a)$ is defined as:

\vspace{-0.1in}
\[
r_{\psi}(s,a) = f\bigl(\psi, c_1(s,a), ..., c_k(s,a)\bigr), ~~~~\psi \in \Psi.
\]
\vspace{-0.1in}

Above, $c_1, ..., c_k$ denote the $k$ reward components, $\psi$ denotes the specification , and $\Psi$ denotes the considered specifications given a function class $f$. Single-task RL typically optimizes a fixed specification $\psi^\star$, whereas multi-objective and multi-task learning can be viewed as considering a set of reward specifications corresponding to different preferences, objectives, or tasks~\citep{roijers2013survey}. 

\section{Expanded Related Work}
\label{app:related_works}

\textbf{Auxiliary objectives.}
Auxiliary objectives have been used to improve representation learning and sample efficiency in RL. UNREAL~\citep{jaderberg2016reinforcement} augments the main RL objective with auxiliary tasks such as pixel control and reward prediction, while methods such as DeepMDP~\citep{gelada2019deepmdp} and SPR~\citep{schwarzer2020data} use objectives over latent dynamics to improve data efficiency. RCRL similarly uses additional supervision to improve a primary task, but the additional objectives are alternative reward specifications on which the agent is conditioned. In consequence, diverse reward supervision can improve nominal task learning while also producing reward-specific behaviors that remain accessible at deployment.

\textbf{Return and Command-Conditioned Policies.}
A related line of work conditions policies on desired outcomes, such as target returns, horizons, or behavioral commands. UDRL~\citep{schmidhuber2019reinforcement} trains policies to predict actions conditioned on commands specifying desired return and time horizon, while reward-conditioned policies and Decision Transformer condition behavior on desired return-to-go~\citep{kumar2019reward, chen2021decision}. These methods enable behavior to be steered under a fixed reward function by changing the desired outcome provided to the policy. RCRL differs in that the conditioning variable is not a target return or command, but the reward specification itself. Thus, changing the conditioning in RCRL changes how rewards are computed, rather than specifying a desired level of return under a fixed reward.

\textbf{Reward-free, unsupervised, and meta-RL.}
Reward-free RL aims to collect experience that supports adaptation to objectives revealed after exploration~\citep{jin2020reward,wang2020reward,wagenmaker2022reward}. Related approaches learn broadly reusable behaviors or representations through unsupervised pretraining~\citep{frans2024unsupervised,ingebrand2024zero,kim2408unsupervised}, model-based exploration~\citep{sekar2020planning}, or pretrained behavioral representations~\citep{sikchi2025fast}. Meta-RL methods, such as VariBAD~\citep{zintgraf2021varibad}, instead adapt to latent tasks by inferring task information from interaction history. RCRL targets a different setting, focusing on the nominal objective and a structured family of alternative rewards known during training, and the reward specification is explicitly provided rather than inferred. Instead of optimizing for broad downstream adaptation, RCRL uses counterfactual reward supervision to improve online learning of the nominal task, with within-family steerability emerging as a secondary capability resulting from off-policy learning of diverse reward functions.

\textbf{Domain randomization and data augmentation.}
Domain randomization and observation augmentation improve robustness by exposing agents to nuisance variation in dynamics or observations~\citep{tobin2017domain,peng2018sim,laskin2020reinforcement,yarats2020image}. These methods typically encourage invariance to the randomized factors. RCRL can instead be viewed as reward augmentation: the same experience is relabeled under different reward specifications, and the agent is explicitly conditioned on that variation. Rather than learning invariance, RCRL preserves reward-specific behavior while using the additional supervision to improve the nominal objective.

\section{Additional Considerations on Training Stability}
\label{app:training_stability}

Training across multiple reward specifications can make critic optimization harder when rewards differ substantially in scale or difficulty. With a standard MSE critic, larger reward scales induce larger TD errors and gradients, which can dominate learning of the shared value function~\citep{hessel2019multi,nauman2025bigger}. In our experiments, most reward specifications remain within a controlled range around the nominal reward, and we do not observe instability beyond that of the underlying backbone. Modern RL algorithms already include mechanisms that reduce this sensitivity. In particular, reward normalization and categorical value learning have been shown to stabilize learning across heterogeneous rewards~\citep{andrychowicz2021matters,bellemare2017distributional,farebrother2024stop,nauman2025bigger}. Backbones such as \textsc{SimbaV2} and \textsc{BRC} therefore require no additional stabilization when combined with RCRL. As discussed in Section~\ref{sec:what_makes_rcrl_effective}, model capacity provides a complementary consideration. RCRL asks a shared critic to represent value functions for multiple reward specifications, and our ablations show substantially larger gains with larger networks. This suggests that sufficient critic capacity is important for exploiting the additional supervision provided by RCRL, consistent with related findings in multi-task RL~\citep{nauman2025bigger}.

\section{Implementation Details}
\label{app:implementation_details}

In this section, we describe how RCRL is integrated into each baseline method, and detail the two proposed RCRL implementation procedures (PRC and ARC). 

\subsection{Algorithms}

We briefly summarize the algorithmic integrations considered in this work. Unless otherwise stated, all architectural choices, optimization settings, and training procedures follow the corresponding baseline implementations, with modifications limited to the incorporation of reward conditioning. 

\textbf{SimbaV2+RCRL and BRC+RCRL.} The integration of RCRL into \textsc{SimbaV2} and \textsc{BRC} follows the same high-level procedure, as both backbones use reward normalization and categorical value learning. Both algorithms normalize rewards and learn categorical value distributions over bounded return ranges $[V_{\min}, V_{\max}]$, which requires care when training with multiple rewards. In principle, one could design reward specifications whose resulting rewards are already normalized. Instead, we use a simpler strategy that avoids additional reward engineering: we maintain separate running normalization statistics for each reward specification used during training. Since normalization statistics are tracked per specification, we instantiate PRC using a finite set of reward specification. Specifically, at the beginning of training, we sample $1024$ perturbation vectors using the log-uniform procedure described in Section~\ref{sec:experiments}, and keep this set fixed throughout training. ARC is finite by construction, since it uses the reward functions associated with the available auxiliary tasks. Because both PRC and ARC therefore operate over finite specification sets in these experiments, we use the embedding-based conditioning strategy described in Section~\ref{sec:method}. Each reward specification is assigned a learnable embedding, which is trained jointly with the critic through the critic objective and used to condition both the critic and policy. For \textsc{BRC}, we reuse its existing task-conditioning machinery, treating the active reward specification as the conditioning identity.

\textbf{DrQV2+RCRL.} In contrast to \textsc{SimbaV2} and \textsc{BRC}, \textsc{DrQV2} uses a standard MSE critic loss and does not employ reward normalization, which makes the integration of RCRL simpler. We therefore apply reward-conditioned training directly with the MSE critic loss, without adding normalization or distributional value learning. This allows us to evaluate whether RCRL remains stable without the stabilization mechanisms used by the other base algorithms. In this setting, reward specifications are sampled from a continuous space, and the corresponding perturbation vectors are used directly for conditioning by concatenating them to the agent input, without a separate embedding module.

\subsection{Conditioning}

In this subsection, we describe the implementation of the two strategies used to generate the set of alternative reward specifications in RCRL. We focus on how the set $\Psi$ is instantiated in practice, how reward specifications are sampled during training, and how conditioning information is incorporated.

\begin{figure}[!ht]
  \begin{center}
  \vspace{-0.1in}
\centerline{\includegraphics[width=0.95\linewidth]{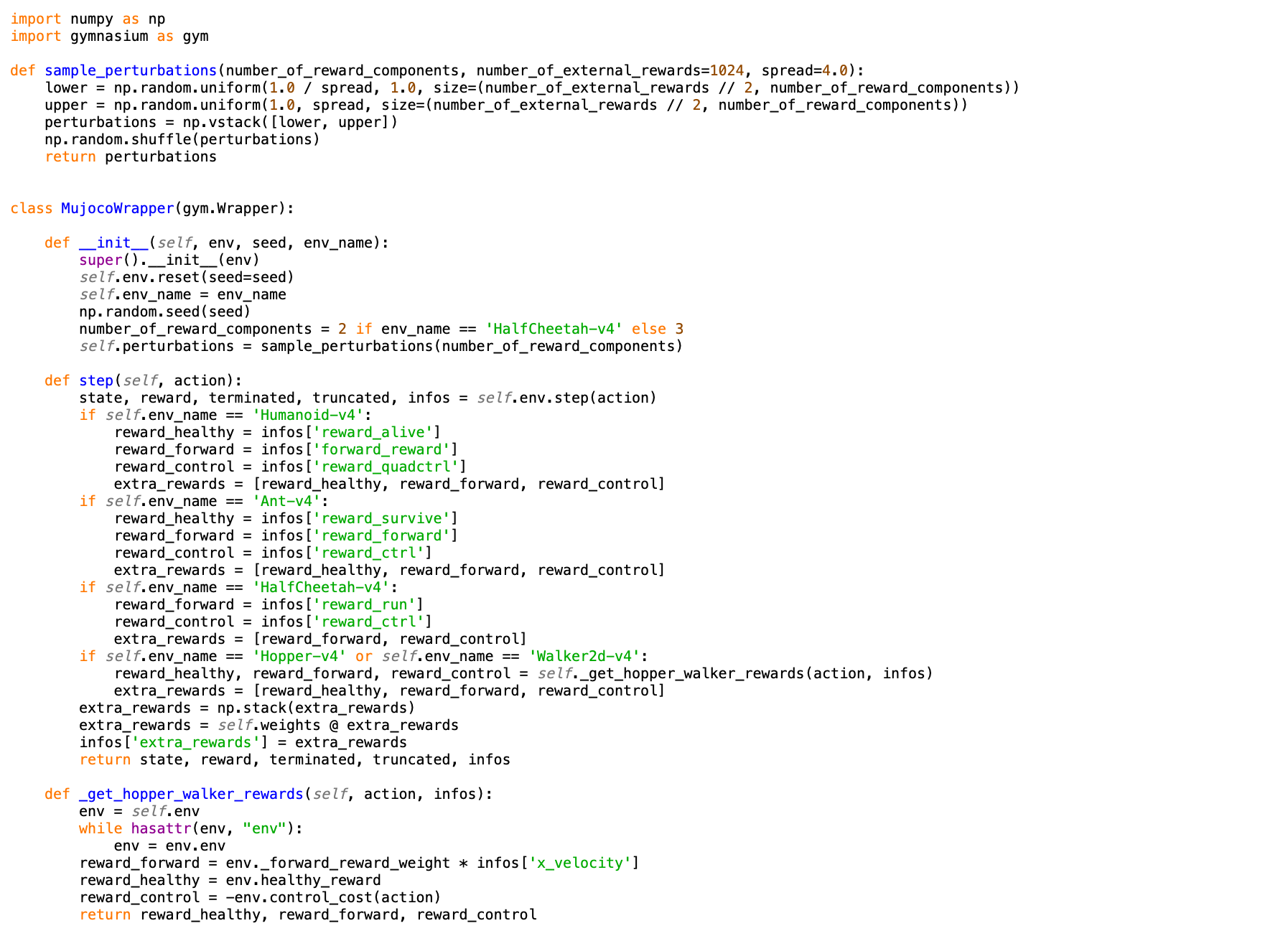}}
    \caption{\footnotesize \emph{\textbf{Code example.}} 
    We present the code used to sample perturbations vector $\Delta$, as well as an OpenAI Gym wrapper used to calculate alternative reward functions in our single-task experiments.}
\label{fig:codesnippet1}
  \end{center}
\vspace{-0.1in}
\end{figure}

\textbf{Perturbed Reward Conditioning (PRC).} For PRC, the set $\Psi$ is defined implicitly as a continuous family of reward specifications obtained by multiplicatively perturbing the nominal reward coefficients. For \textsc{SimbaV2} and \textsc{BRC}, we first sample a finite collection of $1024$ perturbation vectors from the continuous PRC distribution. During training, alternative specifications are sampled independently for each transition from this fixed collection. For \textsc{DrQV2}, perturbations are instead sampled directly from the continuous distribution during training. We present an implementation of PRC in Figure~\ref{fig:codesnippet1}. 

In practice, the benchmarks used in our experiments (i.e. DM Control, OpenAI Gym, and HumanoidBench) employ reward functions that are either linear combinations of reward components or multiplicative (or exponentiated) compositions. For linear reward functions of the form $r_{\psi}(s,a) = \sum_{i=1}^{k} \psi_i~ c_i(s,a)$ the perturbations are applied directly to the linear coefficients $\psi_i$. For multiplicative reward functions of the form $r_{\psi}(s,a) = \prod_{i=1}^{k} c_i(s,a)^{\psi_i}$ the perturbations are applied to the corresponding exponents. This unified specification allows the same perturbation mechanism to be applied consistently across different reward structures. 

\begin{figure}[!ht]
  \begin{center}
  \vspace{-0.1in}
\centerline{\includegraphics[width=0.95\linewidth]{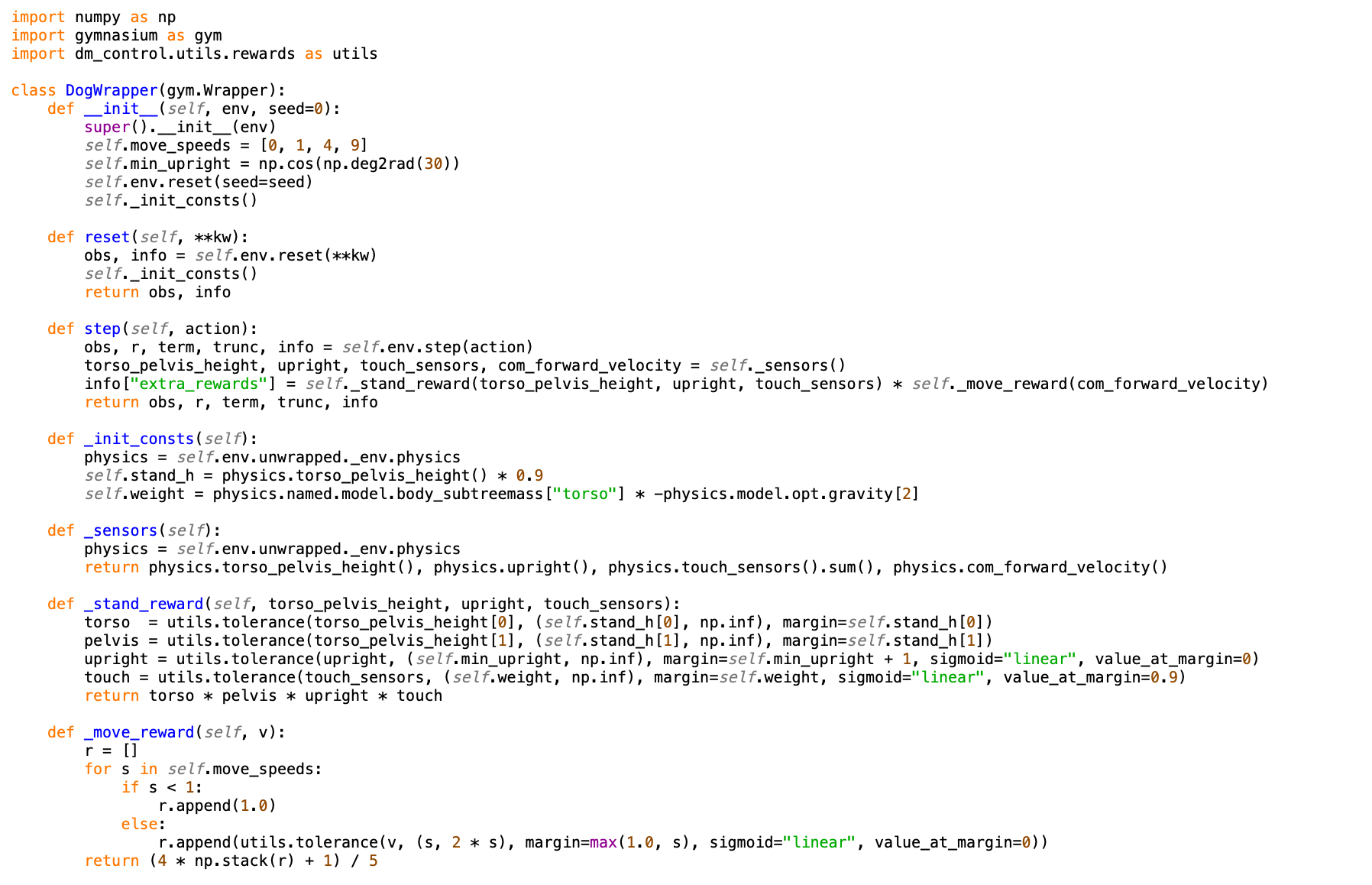}}
    \caption{\footnotesize \emph{\textbf{Code example.}} 
    We present the code used to calculate auxiliary task rewards for the DeepMind Control \texttt{dog} tasks. We use this wrapper in our multi-task experiments.}
\label{fig:codesnippet2}
  \end{center}
\vspace{-0.1in}
\end{figure}

\textbf{Auxiliary Reward Conditioning (ARC).} For ARC, the set $\Psi$ is instantiated explicitly as a finite collection of task-specific reward functions defined within each benchmark. Specifically, auxiliary task rewards are constructed by evaluating the reward functions of other tasks that share the same agent embodiment on the collected state-action trajectories. For example, in the DeepMind Control Suite dog benchmarks, tasks such as \texttt{dog-run}, \texttt{dog-walk}, and \texttt{dog-trot} share an identical reward structure that differs only in the target velocity of the dog. We therefore implement auxiliary reward conditioning by evaluating all reward functions on every transition collected under the nominal task, yielding multiple reward signals corresponding to different target speeds. We show our implementation of this procedure in Figure~\ref{fig:codesnippet2}.

In contrast, in the HumanoidBench benchmark, reward functions are often environment-specific rather than task-specific. For instance, tasks such as \texttt{stair} and \texttt{slide} use identical reward definitions but differ only in the environment layout (e.g. a staircase versus a sliding surface). Moreover, in these tasks the state representation consists solely of proprioceptive features (e.g. joint positions and velocities) and does not encode information about external objects in the environment. As a result, evaluating certain task reward functions on trajectories collected in other environments can yield identical reward signals. To avoid redundancy, we construct $\Psi$ by retaining only reward functions that produce distinct reward outputs on the collected data, discarding any auxiliary reward functions that are equivalent to others. Consequently, the number of distinct auxiliary rewards may be smaller than the number of tasks in the benchmark. We list the exact rewards used in each benchmark in Table~\ref{tab:auxiliary_tasks}.

\begin{table}[b!]
\centering
\footnotesize
\setlength{\tabcolsep}{10pt}
\caption{\footnotesize \emph{\textbf{ARC rewards.}} Tasks in parentheses share identical reward functions and are treated as a single reward specification.}
\vspace{5pt}
\label{tab:auxiliary_tasks}
\begin{tabular}{lll}
\toprule
\textbf{DMC Dogs} & \textbf{HumanoidBench} & \textbf{HumanoidBench Hard} \\
\midrule
\texttt{stand} & \texttt{crawl} & \texttt{crawl} \\
\texttt{walk}  & \texttt{stand} & \texttt{stand} \\
\texttt{trot}  & \texttt{walk}  & \texttt{walk} \\
\texttt{run}   & \texttt{run}   & \texttt{run} \\
               & (\texttt{stair}, \texttt{slide}) & (\texttt{stair}, \texttt{slide}) \\
               & \texttt{pole}  & \texttt{pole} \\
               & \texttt{hurdle} & \texttt{hurdle} \\
               &                & (\texttt{balance\_simple}, \texttt{balance\_hard}) \\
               &                & (\texttt{sit\_simple}, \texttt{sit\_hard}) \\
\bottomrule
\end{tabular}
\end{table}

\section{Experimental Details}
\label{app:experimental_details}

Here, we discuss the details relevant to reproduction of our experiments. All experiments were run using NVIDIA A100 GPU with 40GB of RAM and 8 CPU cores of AMD EPYC 7742.

\subsection{Benchmarks}
\label{app:benchmarks}

In the single-task setting, where RCRL augments \textsc{SimbaV2}, we adopt a subset tasks used in \citet{lee2025hyperspherical}. In the multi-task setting, we use the benchmarks proposed in \citet{nauman2025bigger}.

\textbf{DMC Dogs} \\
\texttt{\small dog-stand, dog-walk, dog-trot, dog-run}

\textbf{OpenAI Gym} \\ 
\texttt{\small HalfCheetah-v4, Humanoid-v4, Hopper-v4, Walker2d-v4, Ant-v4}

\textbf{HumanoidBench} \\ 
\texttt{\small h1-walk, h1-stand, h1-run, h1-stair, h1-crawl, h1-hurdle, h1-slide, h1-pole, h1-maze, h1-reach, h1-sit\_simple, h1-sit\_hard, h1-balance\_simple, h1-balance\_hard}

\textbf{HumanoidBench MT} \\ 
\texttt{\small h1-walk, h1-stand, h1-run, h1-stair, h1-crawl, h1-pole, h1-slide, h1-hurdle, h1-maze}

\textbf{HumanoidBench Hard MT} \\
\texttt{\small h1hands-walk, h1hands-stand, h1hands-run, h1hands-stair, h1hands-crawl, h1hands-pole, h1hands-slide, h1hands-hurdle, h1hands-maze, h1hands-sit\_simple, h1hands-sit\_hard, h1hands-balance\_simple, h1hands-balance\_hard, h1hands-reach, h1hands-spoon, h1hands-window, h1hands-insert\_small, h1hands-insert\_normal, h1hands-bookshelf\_simple, h1hands-bookshelf\_hard}

\textbf{DMC Vision} \\
\texttt{\small acrobot-swingup, cartpole-swingup\_sparse, cheetah-run, hopper-hop, reacher-easy, reacher-hard, quadruped-run, quadruped-walk, walker-run}

\begin{table}[ht]
    \centering
    \caption{\footnotesize \emph{\textbf{Benchmark summary.}} Checkmarks indicate the environment suites included
    in each benchmark.}
    \vspace{5pt}
    \label{tab:benchmarks}
    \small
    \begin{tabular}{lcccc}
        \toprule
        Environment suite & Tasks
        & SimbaV2 & BRC & DrQV2 \\
        \midrule
        DMC Dogs               & 4  & $\checkmark$ & $\checkmark$ & -- \\
        OpenAI Gym              & 5  & $\checkmark$ & -- & -- \\
        HumanoidBench           & 14 & $\checkmark$ & -- & -- \\
        HumanoidBench MT        & 9  & -- & $\checkmark$ & -- \\
        HumanoidBench Hard MT   & 20 & -- & $\checkmark$ & -- \\
        DMC Vision              & 9  & -- & -- & $\checkmark$ \\
        \midrule
        Total tasks & -- & 23 & 33 & 9 \\
        \bottomrule
    \end{tabular}
\end{table}

\subsection{Additional Details on Experiments}
\label{app:experimental_setup}

Here, we provide additional details on the experiments presented in this paper. All figures show $95$\% bootstrapped confidence intervals calculated using the methodology presented in \citet{agarwal2021precipice}.

\subsubsection{RCRL Improves Learning Efficiency}
\label{app:experimental_setup_nominal_reward}

First, we describe the experiments presented in Section~\ref{sec:nominal_performance}. All experiments in this section follow the experimental settings, benchmarks, and evaluation protocols established in prior work.

\textbf{Single-task RL.} We leverage the experimental setup proposed in \textsc{SimbaV2} manuscript~\citep{lee2025hyperspherical}. We use a subset of $3$ out of $4$ benchmarks: $14$ tasks from HumanoidBench, $5$ tasks from OpenAi Gym, and $4$ tasks from DeepMind Control. We list the specific tasks in Appendix~\ref{app:benchmarks}. We implement RCRL via perturbed reward augmentation as described in Section~\ref{sec:method} and Appendix~\ref{app:implementation_details}. We run $10$ seeds for \textsc{SimbaV2} and \textsc{SimbaV2+RCRL}, basing on the official implementation provided in \texttt{https://github.com/DAVIAN-Robotics/SimbaV2}.

\textbf{Multi-task RL.} For multi-task experiments, we adopt the experimental setup from \textsc{BRC}~\citep{nauman2025bigger}. We evaluate on three benchmarks: DMC Dogs ($4$ tasks), HumanoidBench ($9$ tasks), and HumanoidBench Hard ($20$ tasks). We list all tasks in Appendix~\ref{app:benchmarks}. In this setting, we implement auxiliary reward conditioning as described in Section~\ref{sec:method} and Appendix~\ref{app:implementation_details}, conditioning each task on the reward functions of all other non-redundant rewards sharing the same embodiment. We run $5$ seeds per benchmark and report results following the original \textsc{BRC} evaluation protocol and build on the official implementation provided in \texttt{https://github.com/naumix/BiggerRegularizedCategorical}.

\textbf{Vision-based RL.} For vision-based single-task experiments, we follow the experimental setup introduced in \textsc{DrQV2}~\citep{yarats2021drqv2}. We evaluate on the DMC-medium benchmark and exclude two tasks that use a single reward components (\texttt{finger-turn\_easy} and \texttt{finger-turn\_hard}), as they are incompatible with our reward specification strategy. We run $5$ random seeds for each method and list the evaluated tasks in Appendix~\ref{app:benchmarks}. RCRL is implemented using perturbed reward conditioning as described in Section~\ref{sec:method} and Appendix~\ref{app:implementation_details}, with conditioning performed via a standardized perturbation vector concatenated to the state. We build on the implementation provided in \texttt{https://github.com/sukhijab/maxinforl\_jax}.

\subsubsection{What Makes RCRL Effective?}
\label{app:experimental_setup_ablation}

Here, we provide additional details on the ablation experiments presented in Section~\ref{sec:what_makes_rcrl_effective}. All experiments use $10$ random seeds unless directly stated differently.

\begin{table}[ht]
    \centering
    \caption{\footnotesize \emph{\textbf{Effect of batch size.}}
    Comparison of \textsc{SimbaV2} and \textsc{SimbaV2+RCRL} on $14$ HumanoidBench tasks
    after $500$k environment steps, averaged over $10$ seeds.
    Gain denotes the relative improvement of RCRL over SimbaV2.}
    \vspace{5pt}
    \label{tab:batch_size}
    \small
    \begin{tabular}{lccc}
        \toprule
        Total batch size & \textsc{SimbaV2} & \textsc{SimbaV2+RCRL} & RCRL Gain \\
        \midrule
        256 & $0.33 \pm 0.003$ & $0.36 \pm 0.006$ & $+9.1\%$ \\
        512 & $0.32 \pm 0.006$ & $0.37 \pm 0.004$ & $+15.6\%$ \\
        \bottomrule
    \end{tabular}
\end{table}

\begin{figure}[t]
  \begin{center}
  
  \begin{subfigure}{0.32\linewidth}
    \centering
    \includegraphics[width=0.92\linewidth]{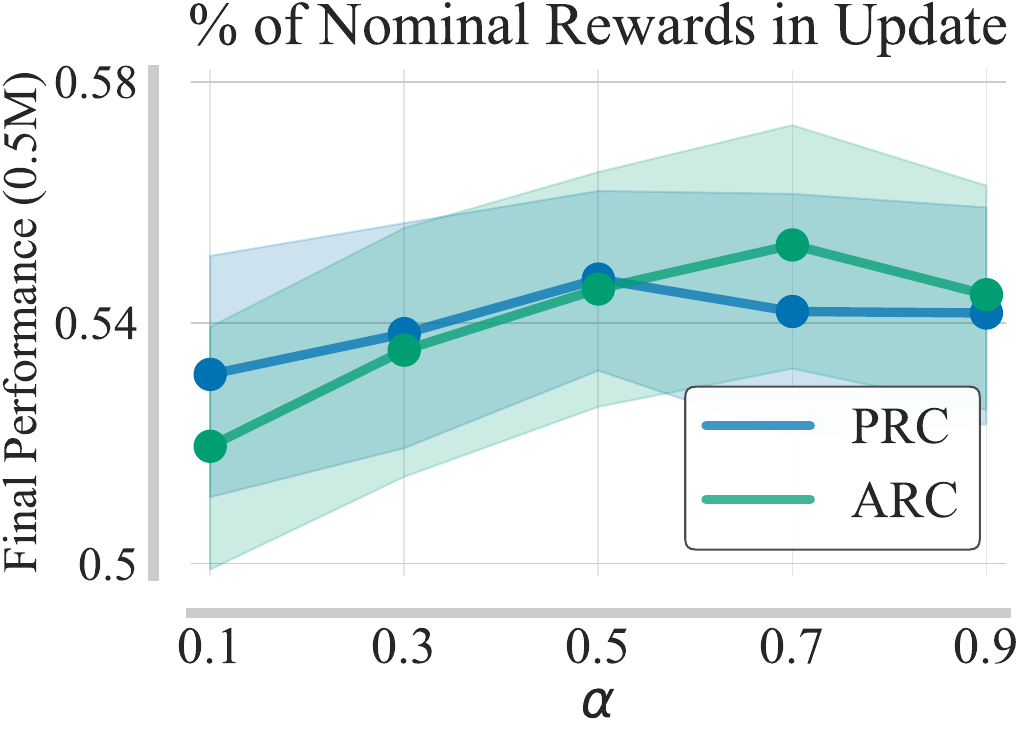}
    \end{subfigure}
    \hfill
  \begin{subfigure}{0.32\linewidth}
    \centering
    \includegraphics[width=0.92\linewidth]{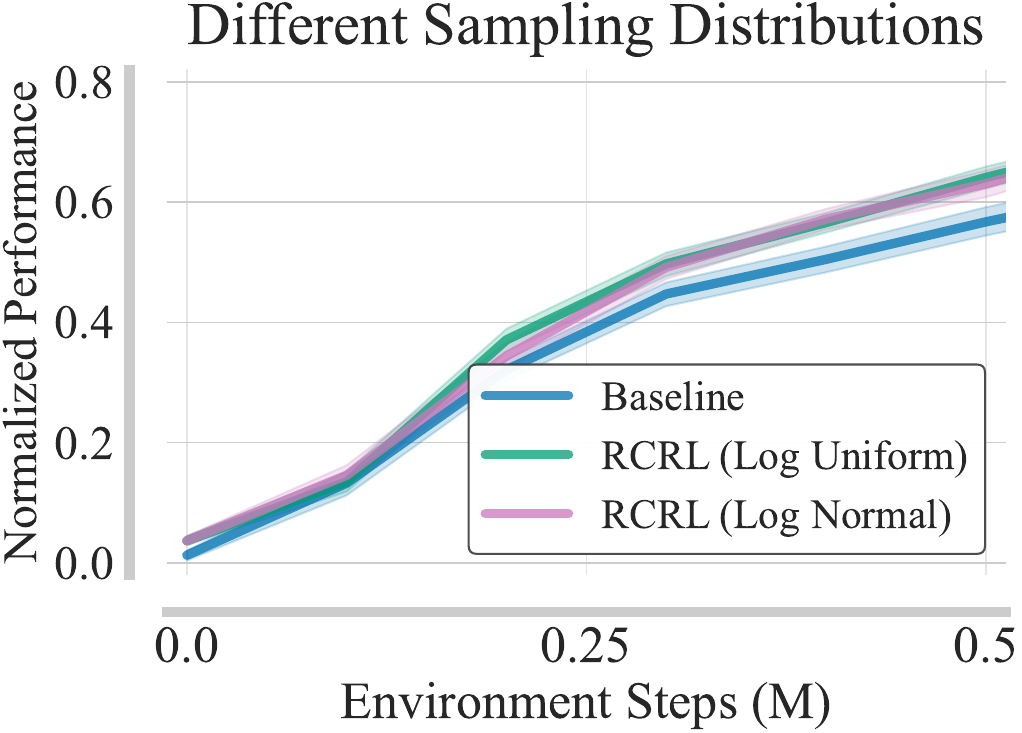}
    \end{subfigure}
    \hfill
  \begin{subfigure}{0.32\linewidth}
    \centering
    \includegraphics[width=0.98\linewidth]{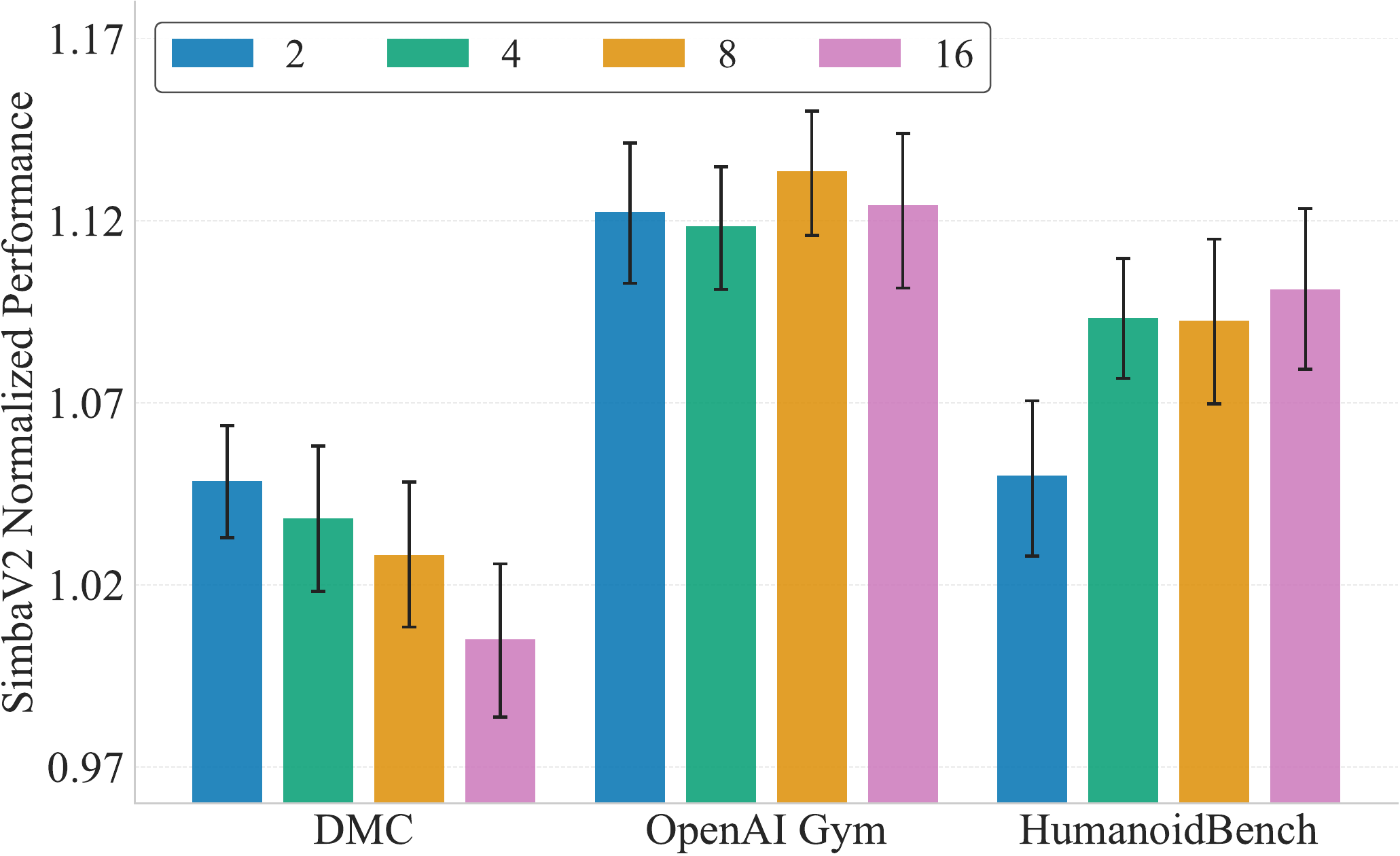}
    \end{subfigure}
    \caption{\footnotesize \emph{\textbf{RCRL sensitivity to counterfactual supervision.}} All experiments use $500$k steps and $10$ seeds. (\textbf{Left}). Sensitivity to the conditioning probability $\alpha$, which controls the fraction of updates performed under the nominal reward. (\textbf{Middle}) Comparison of perturbation sampling distributions for PRC on $14$ HumanoidBench tasks. We compare the default log-uniform to a log-normal centered around the nominal specification. (\textbf{Right}) using the same $14$ tasks, we vary the spread of the log-uniform distribution used to sample perturbations $\text{spread} \in \{2,4,8,16\}$ while keeping all other hyperparameters fixed.}
\label{fig:results_ablation3}
  \end{center}
\end{figure}

\textbf{Reward supervision and coverage.} We use the \textsc{BRC} backbone on the four DMC Dogs tasks, training for $1$M environment steps per task with $5$ seeds. We compare four configurations: Vanilla ST, ST+RCRL, ST+Coverage, and Full MT. For ST+Coverage, we progressively introduce replay transitions collected by single-task agents on the remaining three tasks, following the tandem RL setup~\citep{ostrovski2021difficulty}. This provides $4\times$ more transitions while retaining the nominal reward objective.

\textbf{Reward representation.} We evaluate four conditioning strategies using the ARC variant of \textsc{SimbaV2+RCRL}: no conditioning, where the agent learns a single policy for all reward functions; fixed preference vectors using one-hot encoding; learned reward embeddings following \textsc{BRC}~\citep{nauman2025bigger}; and separate reward-specific heads. We evaluate all variants on $18$ tasks from HumanoidBench and DMC, training for $1$M environment steps with the same reward specifications and training protocol.

\textbf{Reward supervision and exploration.} We compare \textsc{SimbaV2}, RCRL, and RCRL+Explore using both PRC and ARC on the same $18$ HumanoidBench and DMC tasks, training each configuration for $500$k environment steps. RCRL+Explore samples reward specifications for environment interaction from the same distribution used during training, whereas standard RCRL always interacts under the nominal specification. All other settings are kept fixed.

\textbf{Sensitivity to counterfactual supervision.} We evaluate the nominal reward sampling probability $\alpha$ using \textsc{SimbaV2+RCRL} on the same $18$ HumanoidBench and DMC tasks, training for $500$k steps. We consider $\alpha \in \{0.1,0.3,0.5,0.7,0.9\}$, varying the proportion of nominal and counterfactual updates while keeping the reward sampling distribution fixed. The results are presented in Figure~\ref{fig:results_ablation3}.

\textbf{Model capacity.} We compare \textsc{SimbaV2} and \textsc{SimbaV2+RCRL} on the original $23$-task benchmark spanning HumanoidBench ($14$ tasks), OpenAI Gym ($5$ tasks), and DMC ($4$ tasks), training for $1$M environment steps. We use network widths of $128$ ($0.3$M parameters) and $512$ ($2.5$M parameters), with the latter corresponding to the default \textsc{SimbaV2} architecture.

\subsubsection{RCRL and Reward Compatibility}
\label{app:experimental_setup_compatibility}

All experiments in Section~\ref{sec:reward_compatibility} use the ARC variant of \textsc{SimbaV2+RCRL} on the $8$ HumanoidBench tasks listed in Table~\ref{tab:auxiliary_tasks}, training for $1$M environment steps with $10$ random seeds. Since \texttt{stair} and \texttt{slide} share the same reward function, we consider $7$ distinct reward specifications, yielding $48$ nominal-counterfactual reward pairs. For the leave-one-out analysis, we compare RCRL trained with the full reward set against variants with one counterfactual reward removed, keeping the training setup and nominal sampling probability fixed. To study reward similarity, we compute Pearson correlations between nominal and counterfactual reward signals evaluated on the same transitions. For gradient compatibility, we compute critic gradients under both rewards on shared minibatches and measure the frequency of conflicting updates following \citet{yu2020gradient}. Finally, to study gradient consistency, we compare conflict frequencies under vanilla and RCRL training, calculating the average gradient conflict over all $20$ checkpoints and $10$ random seeds, considering both nominal-counterfactual gradients and nominal gradients computed from independently sampled minibatches.

\subsubsection{Steerability Beyond the Nominal Objective}
\label{app:experimental_setup_beyond}

Finally, we describe experiments presented in Section~\ref{sec:steerability}. 

\textbf{Zero-shot steerability.} We evaluate zero-shot adaptation on three continuous control tasks from the DeepMind Control Suite: \texttt{cheetah-run}, \texttt{hopper-hop}, and \texttt{humanoid-walk}. For each task, we run $10$ random seeds. All experiments use the same hyperparameter configurations as in the nominal reward experiments, and we evaluate \textsc{SimbaV2+RCRL}, vanilla \textsc{SimbaV2}, and RCRL+Explore (a variant where counterfactual rewards are also used to guide exploration; in practice this implies an algorithm that works similarly to MTRL/MORL, but with reward relabeling). For each task, we define a set of transformed reward functions that promote specific behavioral variations. 

\begin{figure}[!ht]
  \begin{center}
  \vspace{-0.1in}
\centerline{\includegraphics[width=0.95\linewidth]{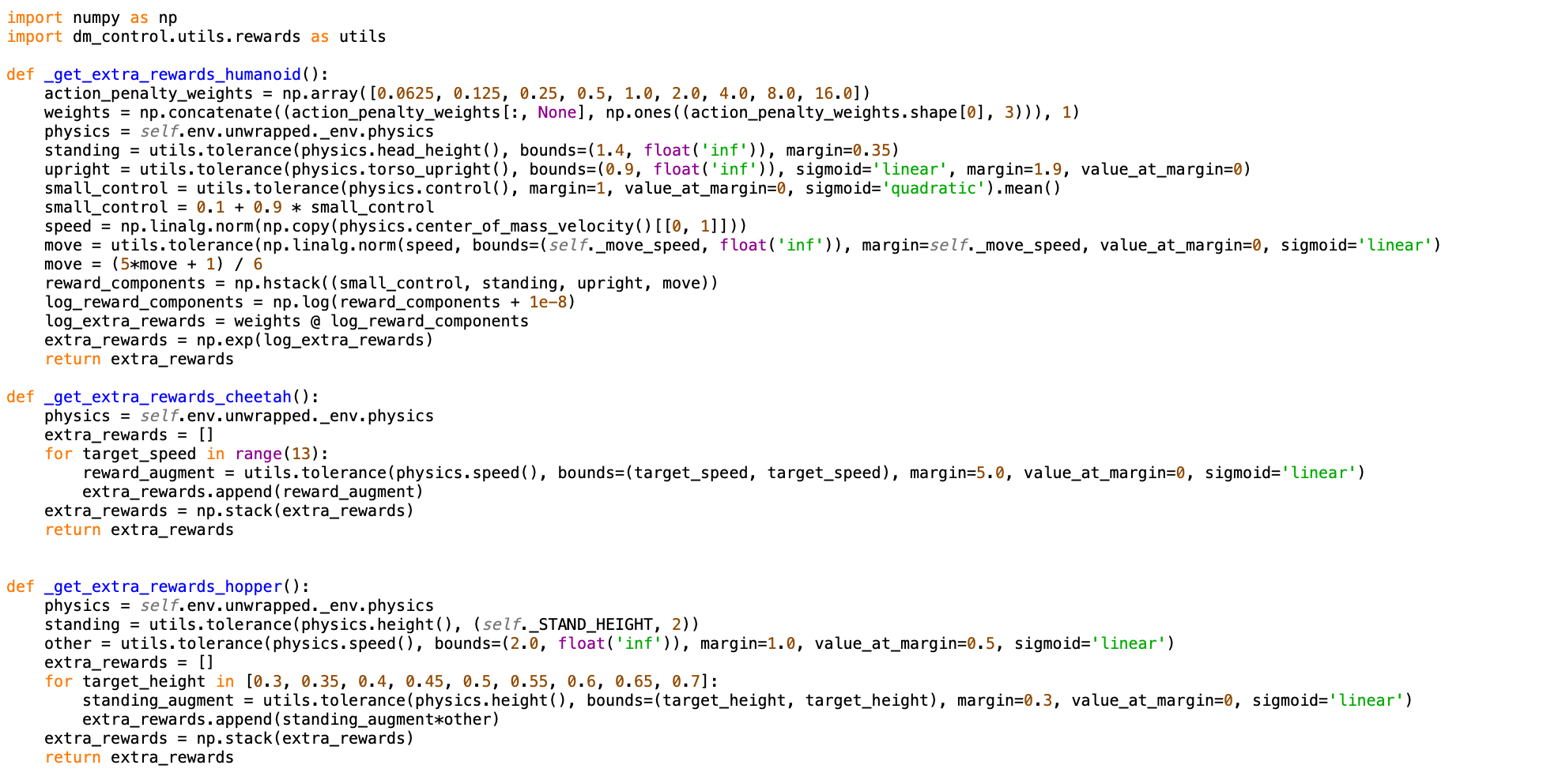}}
    \caption{\footnotesize \emph{\textbf{Code example.}} We present the code used to calculate auxiliary rewards in the zero-shot steerability experiments.}
\label{fig:codesnippet3}
  \end{center}
\vspace{-0.1in}
\end{figure}

For the standard RCRL agent, these transformed rewards are used only for reward conditioning and off-policy updates; the agent does not collect additional experience or explore under these alternative rewards, and all environment interaction is performed exclusively under the nominal reward. After $1$M environment steps of training for \textsc{SimbaV2} and \textsc{SimbaV2+RCRL}, and $3$M environment steps for RCRL+Explore, we evaluate all agents under each of the considered reward functions, reporting the mean return over $10$ rollout episodes per reward specification. For RCRL+Explore, we resample the objective used for exploration after every episode. Zero-shot evaluation is performed by changing only the reward conditioning provided to the policy at test time, without updating any network parameters. Specifically, for \texttt{cheetah-run}, we construct alternative rewards that are maximized when the agent runs at specific target speeds, with desired speeds given by $[0.0, 1.0, 2.0, \ldots, 12.0]$, with maximal recorded speed in single-task training of around $11$. The nominal reward in this task encourages high forward velocity without specifying a target speed. For \texttt{hopper-hop}, auxiliary rewards are defined to favor different target standing heights, with desired heights $[0.30, 0.35, \ldots, 0.70]$. The nominal reward softly enforces a minimum height above $0.6$ but does not prefer any particular height beyond this threshold. Finally, for \texttt{humanoid-walk}, we vary the relative importance of the control cost term by scaling its coefficient with multipliers $[0.0625, 0.125, 0.25, 0.5, 1.0, 2.0, 4.0, 8.0, 16.0]$. 

\textbf{Multi-objective and reward-free methods.} We compare single-task \textsc{SimbaV2}, RCRL, PCC~\citep{yang2019generalized}, USFA~\citep{borsa2018universal}, and FB~\citep{touati2022does} on all $10$ DMC \texttt{walker}, \texttt{dog}, and \texttt{humanoid} tasks. We use a budget of $500$k environment interactions and $5$ random seeds, keeping training settings consistent where possible. We report AUC under both nominal and alternative reward specifications. Since FB is typically trained with substantially more offline data, we evaluate it initialized with replay collected during an RCRL training run from the corresponding seed. 

\begin{table}[ht]
\centering
\caption{\footnotesize \emph{\textbf{Comparison with multi-objective and reward-free methods.}} We report performance aggregated across Walker, Dog, and Humanoid tasks. Each entry shows the AUC on the nominal / alternative objectives.}
\vspace{5pt}
\label{tab:auxiliary_results}
\small
\begin{tabular}{lcccccc}
\toprule
 & Single-task & RCRL & PCC & USFA & FB Online & FB \\
\midrule
Walker   & 790 / 240 & \textbf{828} / 633 & 807 / \textbf{797} & 635 / 588 & 81 / 84 & 642 / 392 \\
Dog      & 420 / 162 & \textbf{441} / \textbf{324} & 238 / 230 & 48 / 49   & 18 / 18 & 95 / 80 \\
Humanoid & 99 / 48   & \textbf{154} / 130 & 138 / \textbf{136} & 3 / 4     & 2 / 2   & 6 / 5 \\
\bottomrule
\end{tabular}
\label{tab:results}
\end{table}

\begin{figure}[t]
  \begin{center}
  \vspace{-0.1in}
\centerline{\includegraphics[width=0.8\linewidth]{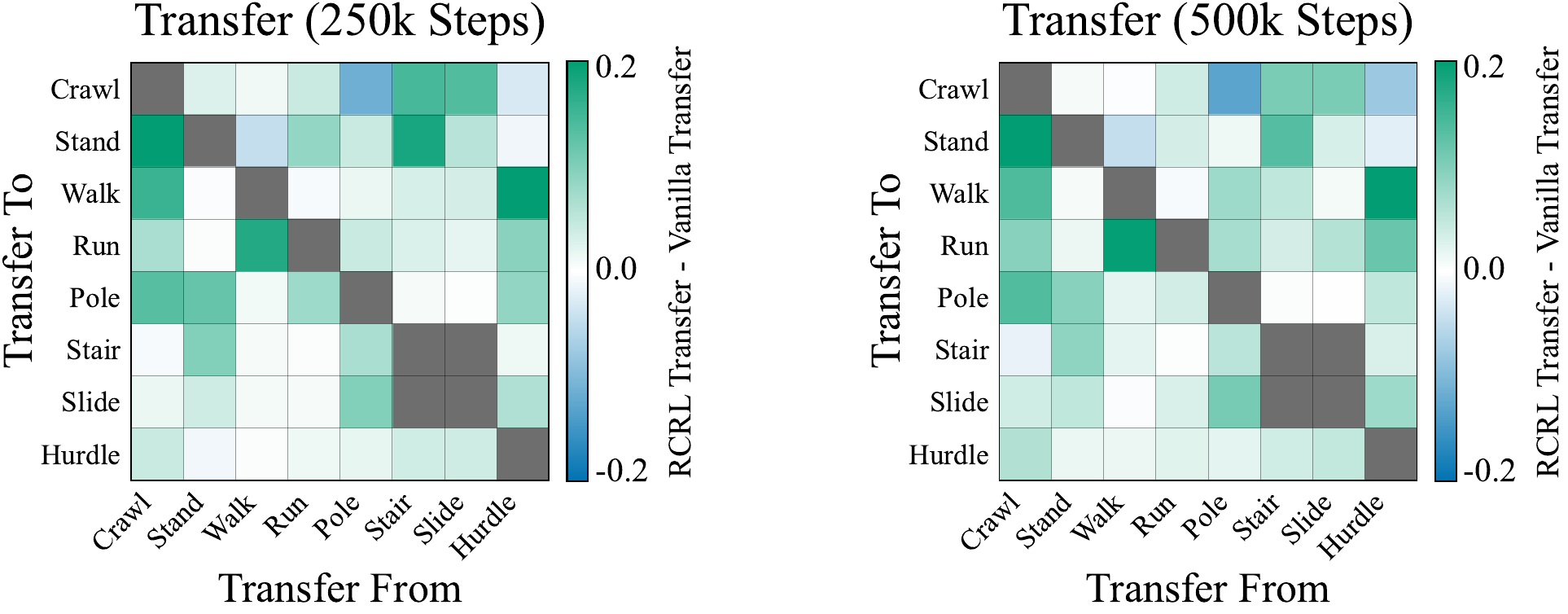}}
\caption{\footnotesize
\emph{\textbf{Transfer to alternative objectives.}}
We compare the transfer performance of \textsc{SimbaV2+RCRL} against vanilla \textsc{SimbaV2} across $54$ source-target pairs from HumanoidBench. The heatmaps show differences in finetuning AUC after $250$k and $500$k target-task steps, with positive values indicating improvements from RCRL. RCRL improves transfer on $74\%$ of pairs, with mean AUC gains of $5$\%. Results are averaged over $10$ random seeds per pair.}
\label{fig:transfer_heatmaps2}
  \end{center}
\vspace{-0.1in}
\end{figure}

\textbf{Transfer to alternative objectives.} We evaluate performance under alternative reward specifications through finetuning experiments on a subset of $8$ tasks from the HumanoidBench benchmark that share identical state and action space structures. The tasks are listed in Table~\ref{tab:auxiliary_tasks}. For each task, we first train an agent for $1$M environment steps under the nominal reward specification using auxiliary reward conditioning. During this phase, the agent is exposed to auxiliary rewards corresponding to the remaining tasks in the set. To evaluate transfer, we initialize a new task by loading the actor and critic parameters from the pretrained source agent, while reinitializing the optimizer state. We then finetune the transferred agent for an additional $500k$ environment steps under the target task reward. Following prior work on online finetuning~\citep{ball2023efficient, zhou2024efficient}, initial exploration in the target environment is performed by sampling actions from the transferred policy rather than using random exploration. For reward-conditioned agents, transfer is implemented by switching the reward embedding used for conditioning from the source task to the embedding corresponding to the target task. This procedure results in $56$ source-target runs, with $10$ seeds per pair. However, since \texttt{stair} and \texttt{slide} share the same reward function and differ only in environment layout, the two corresponding transfer directions do not involve a change in reward specification. We therefore exclude these two pairs from the aggregate transfer statistics, leaving $54$ source-target pairs with distinct reward specifications. This protocol is applied to both \textsc{SimbaV2+RCRL} and the vanilla \textsc{SimbaV2} baseline. 

\subsection{Normalization}
\label{app:normalization}

Here, we describe the normalization of the evaluation score used throughout the paper. We follow the procedures adopted in prior work~\citep{lee2025hyperspherical, nauman2025bigger}. For all benchmarks, we use the following normalization:

\vspace{-0.1in}
\begin{align}
    score^{norm} = \frac{score - score^{random}}{score^{optimal} - score^{random}},
\end{align}
\vspace{-0.1in}

where $score^{norm}$ denotes the normalized evaluation returns, $score$ denotes unnormalized evaluation returns, $score^{optimal}$ denotes the returns achieved by an optimal policy as defined in previous works, and $score^{random}$ denotes the returns achieved by a random policy. 

\begin{table}[ht]
\centering
\caption{Random and optimal scores used for normalization.}
\label{tab:hb_scores}     
\vspace{5pt}
\small               
\begin{tabular}{cccc}
\toprule
\textbf{Benchmark} & \textbf{Task} & \textbf{Random Score} & \textbf{Optimal Score} \\
\midrule
DeepMind Control & dog-stand  & 0.000  & 1000  \\
DeepMind Control & dog-walk  & 0.000  & 1000  \\
DeepMind Control & dog-trot  & 0.000  & 1000  \\
DeepMind Control & dog-run  & 0.000  & 1000  \\
OpenAI Gym & Ant-v4  & -70.288  & 3942  \\
OpenAI Gym & HalfCheetah-v4  & -289.415  & 10574  \\
OpenAI Gym & Hopper-v4  & 18.791 & 3226 \\
OpenAI Gym & Humanoid-v4  & 120.423 & 5165 \\
OpenAI Gym & Walker2d-v4  & 2.791 & 3946 \\
HumanoidBench & balance\_hard-v0     & 10.032  & 800  \\
HumanoidBench & balance\_simple-v0   & 10.170  & 800  \\
HumanoidBench & basketball-v0       & 8.979   & 1200 \\
HumanoidBench & bookshelf\_hard-v0   & 14.848  & 2000 \\
HumanoidBench & bookshelf\_simple-v0 & 16.777  & 2000 \\
HumanoidBench & hurdle-v0           & 2.371   & 700  \\
HumanoidBench & insert\_normal-v0    & 1.673   & 350  \\
HumanoidBench & insert\_small-v0     & 1.653   & 350  \\
HumanoidBench & maze-v0             & 106.233 & 1200 \\
HumanoidBench & pole-v0             & 19.721  & 700  \\
HumanoidBench & reach-v0            & -5024 & 12000\\
HumanoidBench & run-v0              & 1.927   & 700  \\
HumanoidBench & sit\_hard-v0         & 2.477   & 750  \\
HumanoidBench & sit\_simple-v0       & 10.768  & 750  \\
HumanoidBench & slide-v0            & 3.142   & 700  \\
HumanoidBench & spoon-v0            & 4.661   & 650  \\
HumanoidBench & stair-v0            & 3.161   & 700  \\
HumanoidBench & stand-v0            & 11.973  & 800  \\
HumanoidBench & walk-v0             & 2.505   & 700  \\
HumanoidBench & window-v0           & 2.713   & 650  \\
\bottomrule
\vspace{3pt}
\end{tabular}
\end{table}

\subsection{Hyperparameters}
\label{app:hyperparameters}

We detail the hyperparameters used in our experiments in Tables~\ref{tab:sac_brc},~\ref{tab:brc} and~\ref{tab:drq}. As noted in Section~\ref{sec:experiments}, we adopt the configurations of the baseline methods to which RCRL is applied. A detailed discussion of PRC and ARC is provided in Appendix~\ref{app:implementation_details}. Other hyperparameters used in \textsc{SimbaV2}, \textsc{BRC}, \textsc{DrQV2} are in \citet{lee2025hyperspherical}, \citet{nauman2025bigger}, and \citet{yarats2021drqv2} respectively.

\begin{table}[!ht]
  \centering
  \begin{minipage}[t]{0.31\textwidth}
    \caption{SimbaV2+RCRL}
    \label{tab:sac_brc}
    \vspace{5pt}
    \centering
    \small
    \begin{tabular}{l | c }
      \toprule
      \textbf{Hyperparameter} & \textbf{Value} \\ 
      \midrule
      UTD                        & 2    \\ 
      Action repeat              & 2   \\ 
      Discount rate              & Heuristic \\
      Num atoms                  & 101   \\
      $V_{\min}$                 & $-5$ \\
      $V_{\max}$                 & $5$  \\
      Buffer size       & 1e6   \\
      Actor architecture         & SimbaV2  \\
      Actor depth                & 1     \\
      Actor width                & 128   \\
      Critic architecture        & SimbaV2 \\
      Critic depth               & 2     \\
      Critic width               & 512  \\
      Total batch                 & 512  \\
      $\alpha$                 & 0.5  \\
      Conditioning                 & PRC  \\
      \bottomrule
    \end{tabular}
  \end{minipage}%
  \hfill 
  \begin{minipage}[t]{0.31\textwidth}
    \caption{BRC+RCRL}
    \label{tab:brc}
    \vspace{5pt}
    \centering
    \small
    \begin{tabular}{l | c }
      \toprule
      \textbf{Hyperparameter} & \textbf{Value} \\ 
      \midrule
      UTD                        & 2    \\ 
      Action repeat              & 1     \\ 
      Discount rate              & 0.99  \\
      Num atoms                  & 101   \\
      $V_{\min}$                 & $-10$ \\
      $V_{\max}$                 & $10$  \\
      Buffer size per task       & 1e6   \\
      Actor architecture         & BroNet  \\
      Actor depth                & 1     \\
      Actor width                & 256   \\
      Critic architecture        & BroNet \\
      Critic depth               & 2     \\
      Critic width               & 4096  \\
      Total batch                 & 2048  \\
      $\alpha$                 & 0.5  \\
      Conditioning                 & ARC  \\
      \bottomrule
    \end{tabular}
  \end{minipage}
  \hfill 
  \begin{minipage}[t]{0.31\textwidth}
    \caption{DrQV2+RCRL}
    \label{tab:drq}
    \vspace{5pt}
    \centering
    \small
    \begin{tabular}{l | c }
      \toprule
      \textbf{Hyperparameter} & \textbf{Value} \\ 
      \midrule
      UTD                        & 1    \\ 
      Action repeat              & 2     \\ 
      Discount rate              & 0.99  \\
      TD(N)       & 3  \\
      Actor update delay       & 2  \\
      Exploration $\sigma$       & Schedule  \\

      Buffer size        & 1e6   \\
      Actor architecture         & MLP  \\
      Actor depth                & 2     \\
      Actor width                & 256   \\
      Critic architecture        & MLP \\
      Critic depth               & 2     \\
      Critic width               & 256  \\
      Total batch                 & 512  \\
      $\alpha$                 & 0.5  \\
      Conditioning                 & PRC  \\
      \bottomrule
    \end{tabular}
  \end{minipage}
\end{table}

\subsection{Additional Description of Key Figures}
\label{app:figures}

All plots report mean performance across multiple random seeds. Confidence intervals are computed using nonparametric bootstrapping with RLiable~\citep{agarwal2021precipice} and correspond to $95\%$ confidence intervals. 

\textbf{Figures}~\ref{fig:aggregate_results_small} \&~\ref{fig:iclr_timeseries}. Scores are first normalized within each benchmark following the standard practices described in Appendix~\ref{app:normalization}. We then aggregate results across benchmarks by flattening normalized scores across tasks and random seeds. For example, if one benchmark contains 3 tasks and another contains $5$ tasks, and each task is evaluated with $10$ seeds, statistics are computed over an array of shape $[8, 10]$. In both Figures, the performance at each evaluation step is computed by averaging returns over $10$ evaluation rollouts and seeds. We use $10$ seeds for single-task, $5$ seeds for multi-task, $5$ seeds for vision-based, and $10$ seeds for steerability experiments.

\textbf{Figure}~\ref{fig:iclr_heatmap}. This figure presents the reward compatibility analysis across $8$ HumanoidBench tasks, comprising $48$ nominal-counterfactual reward pairs. All experiments use $1$M environment steps and $10$ random seeds. The left panel shows the contribution of individual counterfactual rewards to nominal task learning, measured by the difference in AUC between full RCRL and leave-one-out variants ($C_{i,j}$). The middle panel relates these contributions to the correlation between nominal and counterfactual reward signals, while the right panel examines their relationship with critic gradient conflict frequency calculated according to \citet{yu2020gradient} methodology. 

\textbf{Figure}~\ref{fig:results_zeroshot}. This figure reports zero-shot adaptation results on three DeepMind Control tasks: \texttt{cheetah-run}, \texttt{hopper-hop}, and \texttt{humanoid-walk}. The vanilla single-task agent is trained exclusively under the nominal reward. The \textsc{SimbaV2+RCRL} agent collects experience only under the nominal reward but learns additional reward functions off-policy through reward conditioning. In contrast, the final agent collects experience and learns under all reward functions using relabeling, and has an expanded data budget. All methods are evaluated without further training by varying only the reward specification provided to the policy at test time. For each task, the results are averaged over $10$ random seeds, with evaluation returns computed as the mean over $10$ rollout episodes per reward specification. For each task, the x-axis corresponds to the target behavioral characteristic specified by the reward specification (target speed for \texttt{cheetah-run}, target height for \texttt{hopper-hop}, and control cost multiplier for \texttt{humanoid-walk}). The top row shows the observed behavioral characteristic achieved by the policy at deployment (e.g. actual running speed or standing height), while the bottom row reports the corresponding return under the specified reward function. 

\subsection{Baseline Algorithms}
\label{app:baseline_algorithms}

\textbf{SimbaV2}~\citep{lee2025hyperspherical} is a recent state-of-the-art off-policy actor-critic algorithm for continuous control that combines categorical value learning~\cite{bellemare2017distributional} with hyperspherical~\citep{loshchilov2024ngpt} value representations. It achieves strong performance and sample efficiency across a wide range of locomotion tasks and serves as our primary single-task RL baseline. We combine \textsc{SimbaV2} with RCRL as the backbone for both single-task and transfer experiments. We use the implementation and results provided in the source manuscript (\texttt{https://github.com/DAVIAN-Robotics/SimbaV2}).

\begin{figure}[t]
  \centering
  
  \begin{subfigure}{\linewidth}
    \centering
    \includegraphics[width=0.925\linewidth]{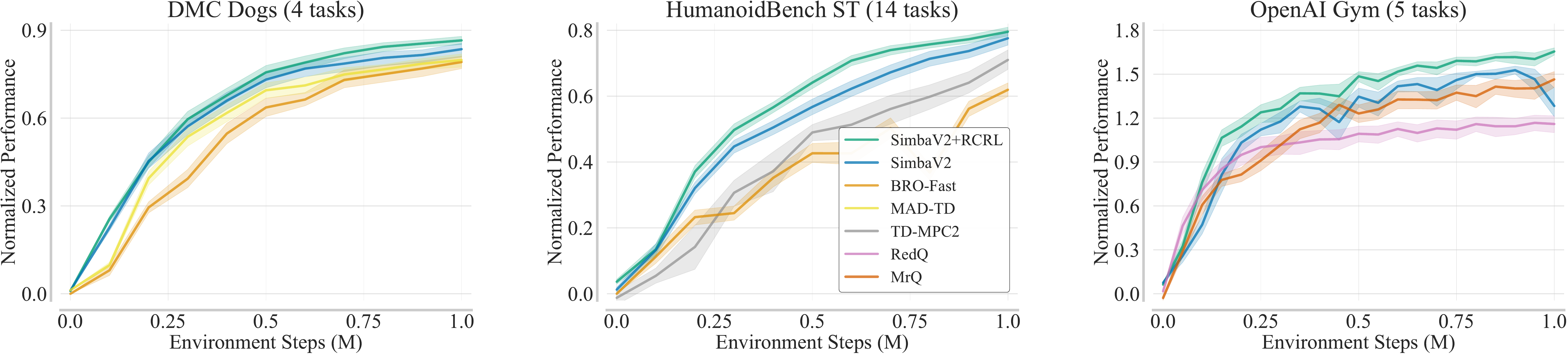}
  \end{subfigure}
  
  \vspace{0.025in}
  
  \begin{subfigure}{\linewidth}
    \centering
    \includegraphics[width=0.925\linewidth]{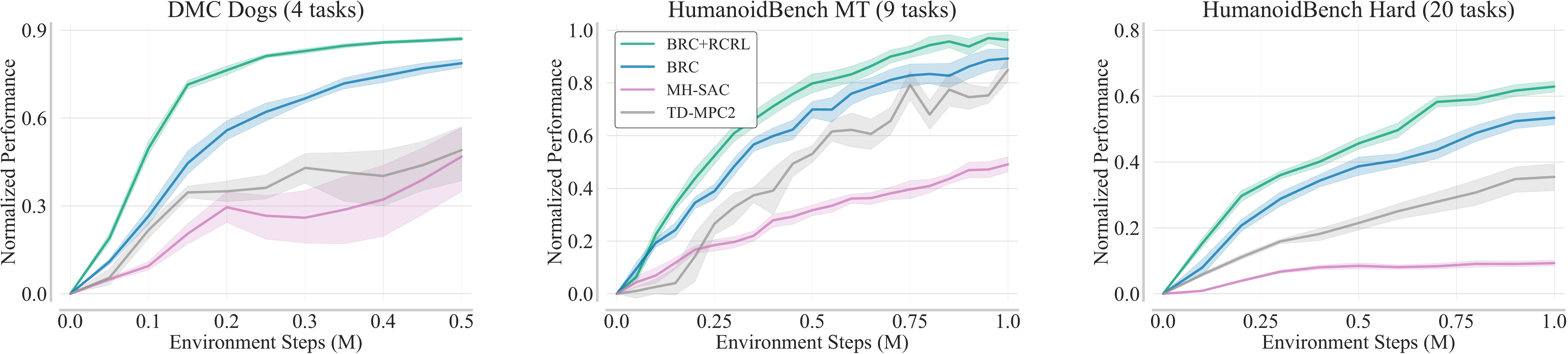}
    \end{subfigure}
    \caption{\footnotesize \emph{\textbf{Sample efficiency of RCRL when evaluated under the nominal reward.}} Results for single-task (top row) and multi-task (bottom row) benchmarks corresponding to results presented in Figure~\ref{fig:iclr_timeseries}.}
  \label{fig:results_timeseries}
  \vspace{-0.2in}
\end{figure}

\textbf{BRC}~\citep{nauman2025bigger} is a single and multi-task RL algorithm designed for efficient learning. Its core design components are scaled value network, conditioning on task embeddings, and distributional value learning with reward normalization. We use \textsc{BRC} as the base model for evaluating RCRL in the multi-task settings. We use the implementation and results provided in the source manuscript (\texttt{https://github.com/naumix/BiggerRegularizedCategorical}).

\textbf{DrQV2}~\citep{yarats2021drqv2} is a strong baseline for vision-based RL that combines data augmentation with off-policy actor-critic learning. It uses relatively simple architectures and optimization techniques compared to recent large-scale methods, making it a useful testbed for evaluating whether RCRL can be applied effectively in pixel-based control without relying on stabilization mechanisms. We use the implementation provided in \texttt{https://github.com/sukhijab/maxinforl\_jax}.

\textbf{BRO}~\citep{nauman2024bigger} is an off-policy reinforcement learning algorithm that demonstrates how scaling the value network can lead to substantial performance improvements in online RL. Its core contribution is a scaled critic architecture based on a normalized ResNet, which enables stable training of large value networks. We use the implementation and results provided in the source manuscript (\texttt{https://github.com/naumix/BiggerRegularizedOptimistic}).

\textbf{MAD-TD}~\citep{voelcker2024mad} is an approach designed to stabilize training under high update-to-data ratios in off-policy RL~\citep{janner2019trust, hiraoka2021dropout, nikishin2022primacy, d2022sample}. It mitigates instability and overestimation by augmenting real replay data with a small amount of model-based transitions. We use the results provided in \citet{lee2025hyperspherical}.

\textbf{TD-MPC2}~\citep{hansen2023tdmpc2} is a model-based RL algorithm that combines latent dynamics modeling with planning and policy learning and was shown to robustly perform across a variety of benchmarks, while using a single hyperparameter configuration. We use the results provided in \citet{lee2025hyperspherical}.

\textbf{REDQ}~\citep{chen2020randomized} is an early approach designed to stabilize training under high update-to-data ratios in off-policy RL by using an ensemble of critics and randomized target selection during updates. It represents a strong and widely used baseline for continuous control in OpenAI gym tasks. We use the results provided in \citet{lee2025hyperspherical}.

\textbf{MrQ}~\citep{fujimoto2025towards} is a model-free RL algorithm that was shown to perform well across both discrete and continuous control tasks using a unified hyperparameter configuration. It integrates model-based representations into a model-free learning process in the style of TD3~\citep{fujimoto2018addressing} to improve robustness across varied tasks. We use the results provided in \citet{lee2025hyperspherical}.

\textbf{MH-SAC}~\citep{yu2019meta} is a multi-head variant of Soft Actor-Critic~\citep{Haarnoja2017, Haarnoja18} designed for multi-task RL, where different tasks are represented via separate output heads while sharing lower-level representations. It provides an early and influential example of conditioning policies on task identity in multi-task RL. We use the implementation and results provided in \citet{nauman2025bigger}.

\section{Training Curves}
\label{app:training_curves}

\begin{figure}[!ht]
  \begin{center}
  
\centerline{\includegraphics[width=0.99\linewidth]{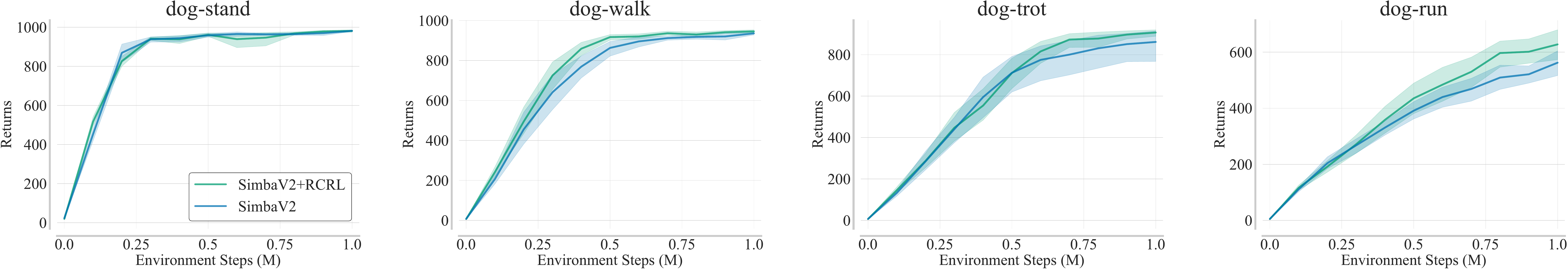}}
    \caption{\footnotesize \emph{\textbf{Single-task DeepMind Control performance.}} 
    We present mean and $95\%$ confidence intervals calculated using bootstrapping. $10$ random seeds.}
\label{fig:tc_single_dmc}
  \end{center}
\vspace{-0.2in}
\end{figure}

\begin{figure}[!ht]
  \begin{center}
  
\centerline{\includegraphics[width=0.99\linewidth]{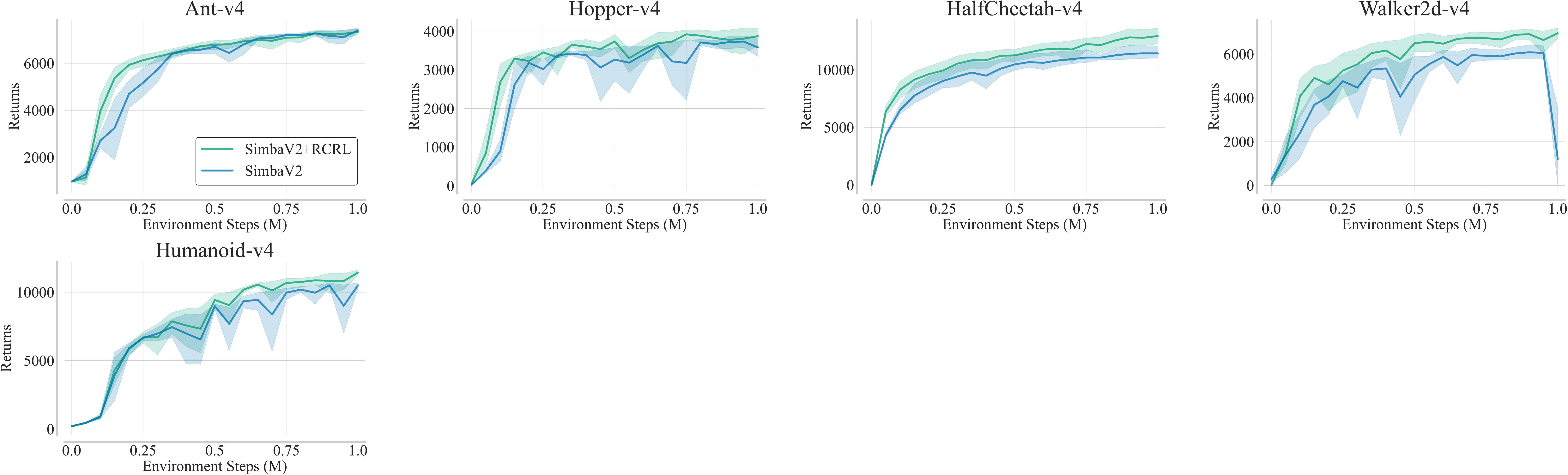}}
    \caption{\footnotesize \emph{\textbf{Single-task Gym performance.}} 
    We present mean and $95\%$ confidence intervals calculated using bootstrapping. $10$ random seeds.}
\label{fig:tc_single_gym}
  \end{center}
\vspace{-0.2in}
\end{figure}

\begin{figure}[!ht]
  \begin{center}
  
\centerline{\includegraphics[width=0.99\linewidth]{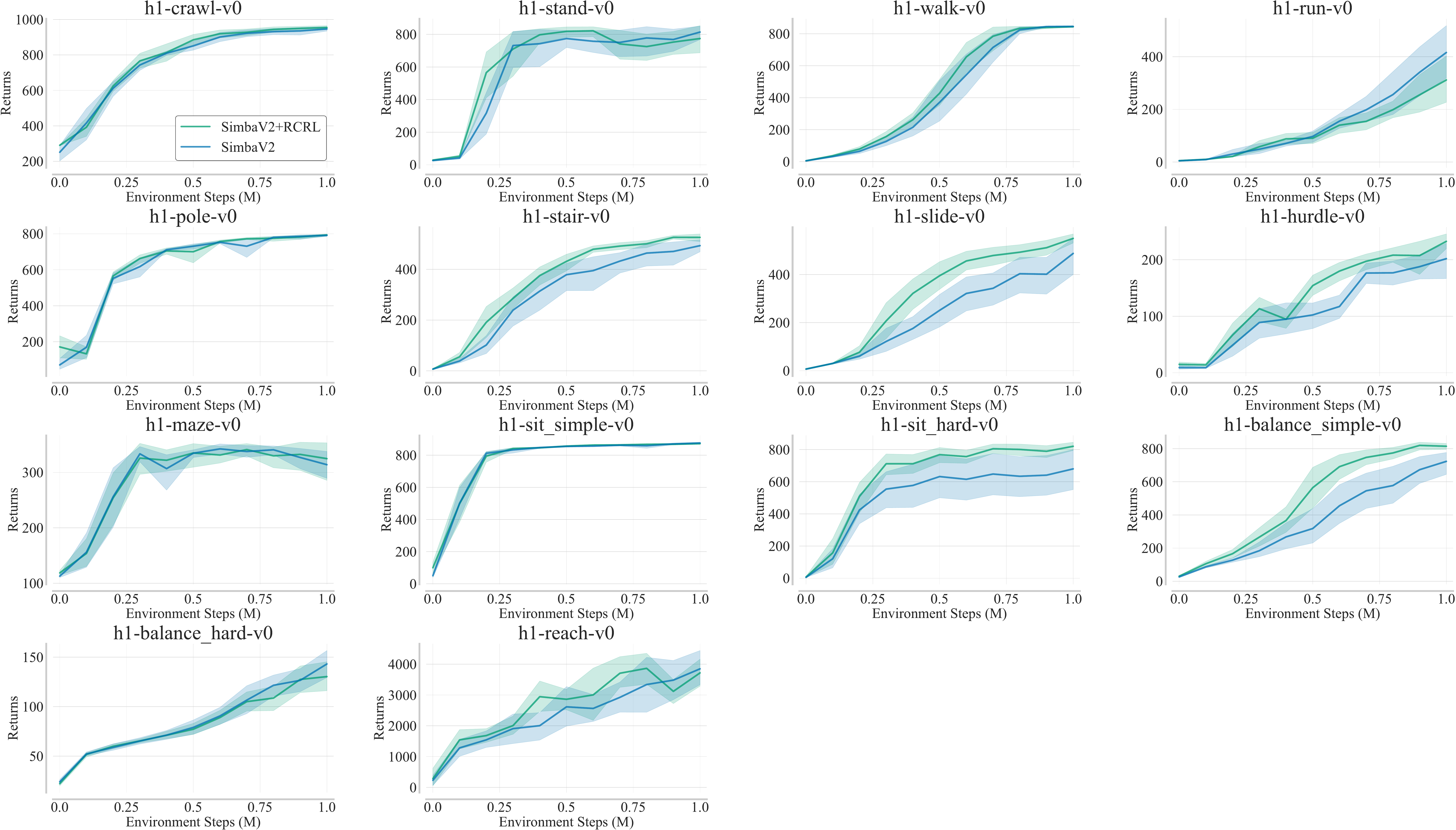}}
    \caption{\footnotesize \emph{\textbf{Single-task HumanoidBench performance.}} 
    We present mean and $95\%$ confidence intervals calculated using bootstrapping. $10$ random seeds.}
\label{fig:tc_single_hb}
  \end{center}
\vspace{-0.2in}
\end{figure}

\begin{figure}[!ht]
  \begin{center}
  
\centerline{\includegraphics[width=0.99\linewidth]{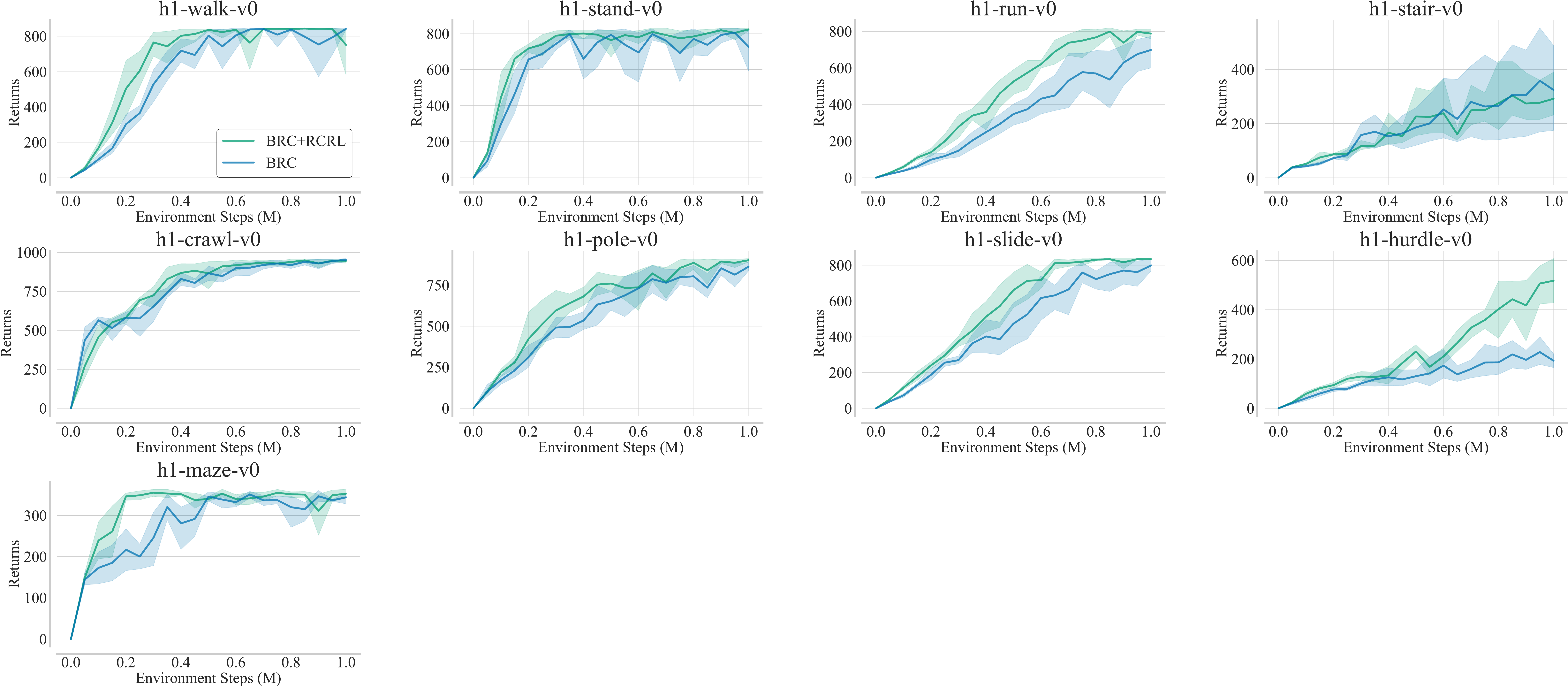}}
    \caption{\footnotesize \emph{\textbf{Multi-task HumanoidBench performance.}} 
    We present mean and $95\%$ confidence intervals calculated using bootstrapping. $5$ random seeds.}
\label{fig:tc_multi_hbe}
  \end{center}
\vspace{-0.2in}
\end{figure}

\begin{figure}[!ht]
  \begin{center}
  
\centerline{\includegraphics[width=0.99\linewidth]{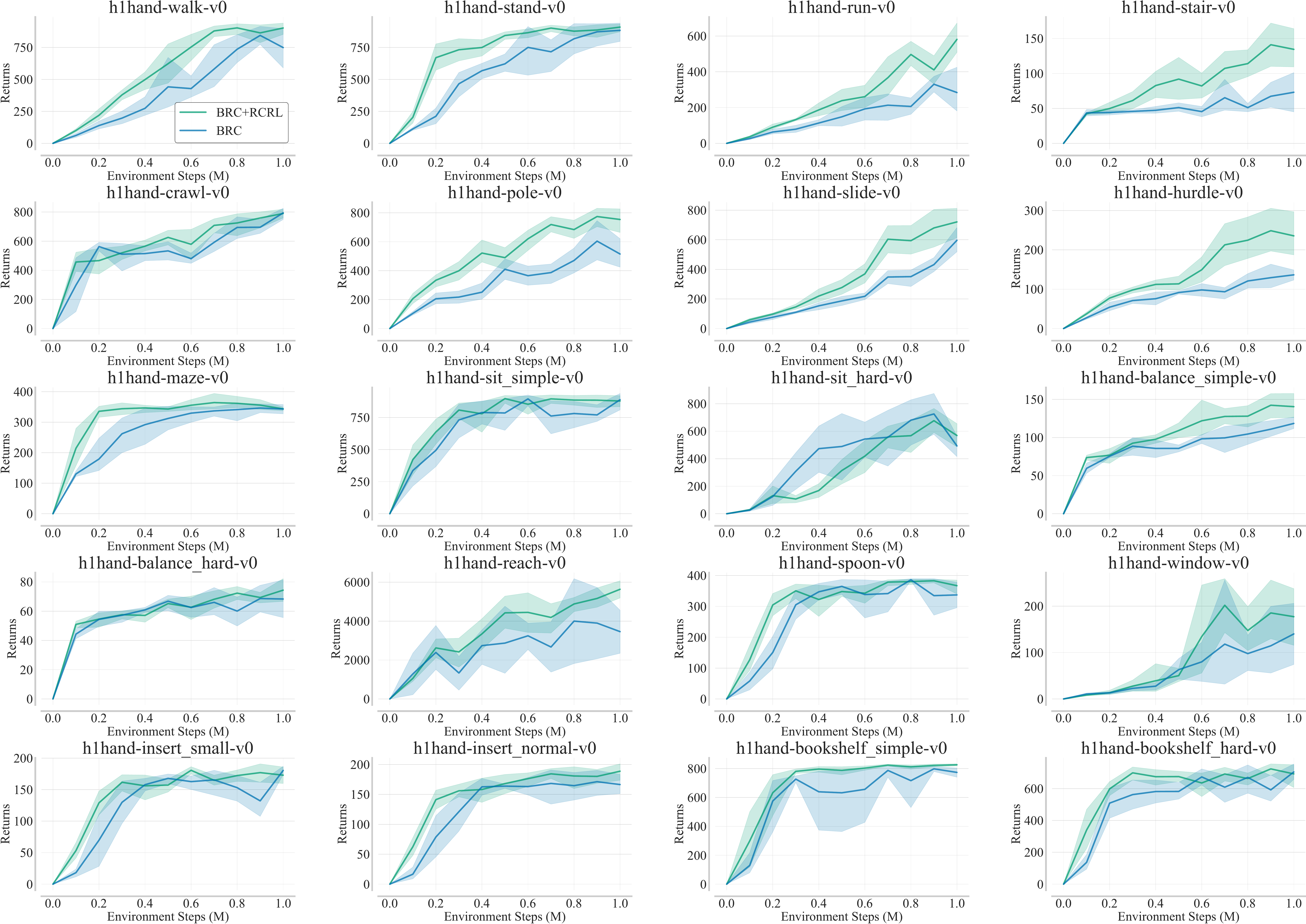}}
    \caption{\footnotesize \emph{\textbf{Multi-task HumanoidBench Hard performance.}} 
    We present mean and $95\%$ confidence intervals calculated using bootstrapping. $5$ random seeds.}
\label{fig:tc_multi_hbh}
  \end{center}
\vspace{-0.2in}
\end{figure}

\begin{figure}[!ht]
  \begin{center}
  
\centerline{\includegraphics[width=0.99\linewidth]{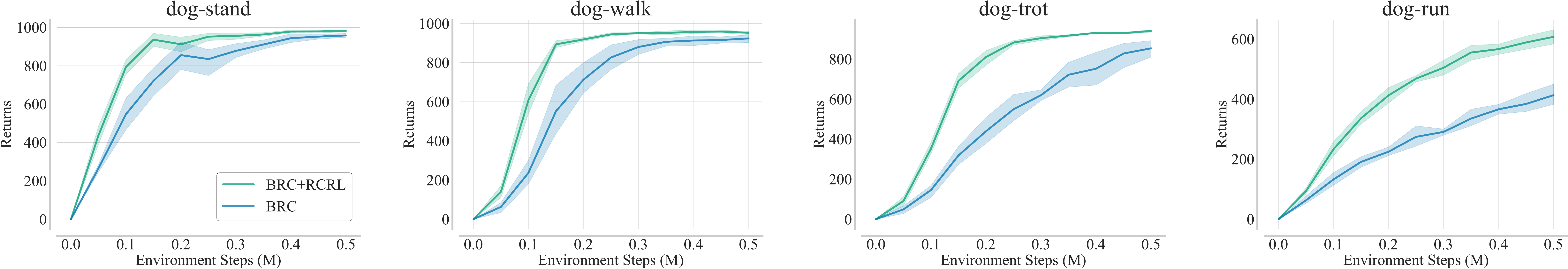}}
    \caption{\footnotesize \emph{\textbf{Multi-task DeepMind Control performance.}} 
    We present mean and $95\%$ confidence intervals calculated using bootstrapping. $5$ random seeds.}
\label{fig:tc_multi_dmc}
  \end{center}
\vspace{-0.2in}
\end{figure}

\begin{figure}[!ht]
  \begin{center}
  
\centerline{\includegraphics[width=0.99\linewidth]{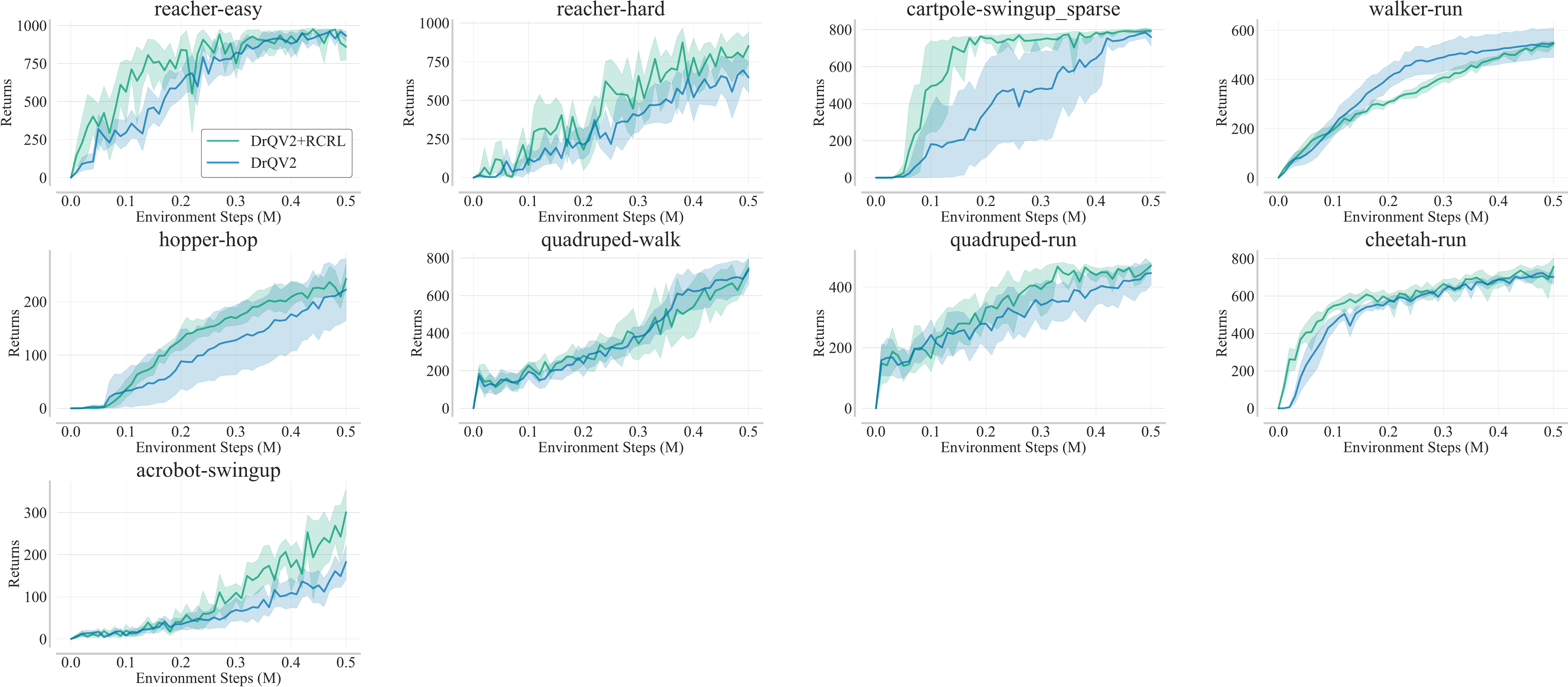}}
    \caption{\footnotesize \emph{\textbf{Vision-based DeepMind Control performance.}} 
    We present mean and $95\%$ confidence intervals calculated using bootstrapping. $5$ random seeds.}
\label{fig:tc_vision}
  \end{center}
\vspace{-0.2in}
\end{figure}

\end{document}